\documentclass[journal, 10]{IEEEtran}
\IEEEoverridecommandlockouts

\ifCLASSINFOpdf
\usepackage[pdftex]{graphicx}
\graphicspath{{/figs/}}
\DeclareGraphicsExtensions{.pdf,.eps}
\else
\usepackage[dvips]{graphicx}
\graphicspath{{/figs/}}
\DeclareGraphicsExtensions{.eps}
\fi

\usepackage{xspace}
\usepackage{caption}
\usepackage{subcaption}
\usepackage[cmex10]{amsmath}
\usepackage{amssymb} 
\usepackage{cite}
\usepackage{bm}
\usepackage{array}

\usepackage{algorithm,tabularx} 
\usepackage[noend]{algpseudocode}
\makeatletter
\newcommand{\multilines}[1]{%
	\begin{tabularx}{\dimexpr\linewidth-\ALG@thistlm}[t]{@{}X@{}}
		#1
	\end{tabularx}
}
\makeatother

\usepackage{mathtools}

\usepackage{amsthm}
\usepackage{multirow}
\usepackage{color}
\usepackage{epstopdf}
\usepackage{endnotes}
\usepackage{framed}
\usepackage{cool} 
\usepackage{enumerate} 
\usepackage[subnum]{cases}

\newcommand{\norm}[1]{\left\lVert#1\right\rVert}

\newcommand{\SumNoLim}[2]{\ensuremath{\sum\nolimits_{#1}^{#2}}}
\newcommand{\SumLim}[2]{\ensuremath{\sum\limits_{#1}^{#2}}}

\newcommand{\defeq}{\mathrel{\mathop:}=}

\newcommand{\innProd}[2]{\ensuremath{\langle{#1}, {#2}\rangle}}

\newcommand{\pder}[2][]{\frac{\partial#1}{\partial#2}} 

\newcommand{\bigP}[1]{\ensuremath{\bigl(#1\bigr)}}

\newcommand{\bigC}[1]{\ensuremath{\bigl\{#1\bigr\}}}
\newcommand{\biggP}[1]{\ensuremath{\biggl(#1\biggr)}}
\newcommand{\biggS}[1]{\ensuremath{\biggl[#1\biggr]}}

\newcommand{\BigP}[1]{\ensuremath{\Bigl(#1\Bigr)}}
\newcommand{\BigS}[1]{\ensuremath{\Bigl[#1\Bigr]}}

\newcommand{\tbf}[1]{\textbf{#1}}

\newcommand{\Cal}[1]{\mathcal{#1}} 
\newcommand{\DD}[1]{\mathbb{#1}} 

\theoremstyle{plain}
\newtheorem{theorem}{Theorem}
\newtheorem{assumption}{Assumption}
\newtheorem{corollary}{Corollary}
\newtheorem{lemma}{Lemma}

\theoremstyle{definition}

\theoremstyle{remark}

\newcommand{\Tcomp}{\ensuremath{T_{cp}\xspace}}
\newcommand{\Tcomm}{\ensuremath{T_{co}\xspace}}
\newcommand{\TcompOpt}{\ensuremath{T^*_{cp}\xspace}}
\newcommand{\TcommOpt}{\ensuremath{T^*_{co}\xspace}}
\newcommand{\TnOne}{\ensuremath{T_{\Cal{N}_1}\xspace}}
\newcommand{\TnTwo}{\ensuremath{T_{\Cal{N}_2}\xspace}}
\newcommand{\TnThree}{\ensuremath{T_{\Cal{N}_3}\xspace}}

\newcommand{\Titer}{\ensuremath{T_{g}\xspace}}
\newcommand{\Eiter}{\ensuremath{E_{g}\xspace}}

\newcommand{\Ecomm}{\ensuremath{E_{n,co}\xspace}}
\newcommand{\Ecomp}{\ensuremath{E_{n,cp}\xspace}}
\newcommand{\EcommOpt}{\ensuremath{E^*_{n,co}\xspace}}
\newcommand{\EcompOpt}{\ensuremath{E^*_{n,cp}\xspace}}

\newcommand{\Lcomm}{\ensuremath{L_{co}\xspace}}
\newcommand{\Lcomp}{\ensuremath{L_{cp}\xspace}}

\newcommand{\Opt}{\textsf{FEDL}\xspace}
\newcommand{\SubOne}{\textsf{SUB1}\xspace}
\newcommand{\SubTwo}{\textsf{SUB2}\xspace}
\newcommand{\SubThree}{\textsf{SUB3}\xspace}

\newcommand{\nab}{\ensuremath{\nabla \bar{F}^{t-1} \xspace}} 
\newcommand{\FL}{\textsf{FEDL}\xspace}

\newcommand{\None}{\ensuremath{\Cal{N}_1\xspace}}
\newcommand{\Ntwo}{\ensuremath{\Cal{N}_2\xspace}}
\newcommand{\Nthree}{\ensuremath{\Cal{N}_3\xspace}}

\newcommand{\fMin}{\ensuremath{f_{n}^{min}\xspace}}
\newcommand{\fMax}{\ensuremath{f_{n}^{max}\xspace}}

\newcommand{\pMin}{\ensuremath{p_{n}^{min}\xspace}}
\newcommand{\pMax}{\ensuremath{p_{n}^{max}\xspace}}

\newcommand{\tauMin}{\ensuremath{\tau_{n}^{min}\xspace}}
\newcommand{\tauMax}{\ensuremath{\tau_n^{max}\xspace}}

\newcommand{\SumN}{\ensuremath{\SumNoLim{n=1}{N}\xspace}}

\def\BibTeX{{\rm B\kern-.05em{\sc i\kern-.025em b}\kern-.08em
		T\kern-.1667em\lower.7ex\hbox{E}\kern-.125emX}}
\begin{document}

	\title{Federated Learning over Wireless Networks: Convergence Analysis and Resource Allocation}

	\author{
		Canh~T.~Dinh,
		Nguyen~H.~Tran, 
		Minh N. H. Nguyen, 
		Choong Seon Hong,
		Wei Bao, 
		Albert Y. Zomaya, 
		Vincent Gramoli
				\IEEEcompsocitemizethanks{
				\IEEEcompsocthanksitem C.~T.~Dinh, N.~H.~Tran, W.~Bao, and A.Y.~Zomaya are with the School of Computer Science, The University of Sydney, Sydney, NSW 2006, Australia (email: tdin6081@uni.sydney.edu.au,\{nguyen.tran, wei.bao, albert.zomaya\}@sydney.edu.au).  V.~Gramoli is with the School of Computer Science, The University of Sydney, Sydney, NSW 2006, Australia and the Distributed Computing Lab at EPFL, Station 14 CH-1015 Lausanne, Switzerland (email: vincent.gramoli@epfl.ch).   M. N. H. Nguyen is with the Department of Computer Science and Engineering, Kyung Hee University, South Korea and also Vietnam - Korea University of Information and Communication Technology, The University of Danang, Vietnam (email: minhnhn@khu.ac.kr). C. S. Hong is with the Department of Computer Science and Engineering, Kyung Hee University, South Korea (email: cshong@khu.ac.kr).	This research is funded by Vietnam National Foundation for Science and Technology Development (NAFOSTED) under grant number 102.02-2019.321. Part of this work was presented at IEEE INFOCOM 2019 \cite{tranFederatedLearningWireless2019}. (Corresponding authors:  Nguyen H. Tran and Choong Seon Hong.) 
			
		}
	}
	
	
	\maketitle
	
	\begin{abstract}
		There is an increasing interest in a fast-growing machine learning technique called Federated Learning (FL), in which the model training is distributed over mobile user equipment (UEs), exploiting UEs' local computation and training data. Despite its advantages such as preserving data privacy, FL still has challenges of heterogeneity across UEs' data and physical resources. To address these challenges, we first propose \FL,  a FL algorithm which can handle heterogeneous UE data without further assumptions except strongly convex and smooth loss functions. We provide a convergence rate characterizing the trade-off between local computation rounds of each UE to update its local model and global communication rounds to update the FL global model. We then employ \FL in wireless networks as a resource allocation optimization problem that captures the trade-off between \FL convergence wall clock time and energy consumption of UEs with heterogeneous computing and power resources. Even though the wireless resource allocation problem of \FL is non-convex, we exploit this problem's structure to decompose it into three sub-problems and analyze their closed-form solutions as well as insights into problem design. Finally, we empirically evaluate the convergence of \FL with PyTorch experiments, and provide extensive numerical results for the wireless resource allocation sub-problems. Experimental results  show that \FL outperforms the vanilla FedAvg algorithm in terms of convergence rate and test accuracy in various settings.

	\end{abstract}
	
	
	\IEEEpeerreviewmaketitle
	
	\section{Introduction} \label{S:Intro}
	
	The significant increase in the number of cutting-edge mobile and Internet of Things (IoT) devices results in the phenomenal growth of the data volume generated at the edge network. It has been predicted that in 2025 there will be 80 billion devices connected to the Internet and the global data will achieve 180 trillion gigabytes \cite{reinselDigitizationWorldEdge2018}. However, most of this data is privacy-sensitive in nature. It is not only risky to store this data in data centers but also costly in terms of communication. For example, location-based services such as the app Waze \cite{FreeCommunitybasedGPS}, can help users avoid heavy-traffic roads and thus reduce congestion. However, to use this application, users have to share their own locations with the server and it cannot guarantee that the location of drivers is kept safely. Besides, in order to suggest the optimal route for drivers, Waze collects a large number of data including every road driven to transfer to the data center. Transferring this amount of data requires a high expense in communication and drivers' devices to be connected to the Internet continuously.
	
	In order to maintain the privacy of consumer data and reduce the communication cost, it is necessary to have an emergence of a new class of machine learning techniques that shifts computation to the edge network where the privacy of data is maintained. One such popular technique is called Federated Learning (FL) \cite{mcmahanCommunicationEfficientLearningDeep2017}. This technology allows users to collaboratively build a shared learning model while preserving all training data on their user equipment (UE). In particular, a UE computes the updates to the current global model on its local training data, which is then aggregated and fed-back by a central server, so that all UEs have access to the same global model to compute their new updates. This process is repeated until an accuracy level of the learning model is reached. In this way, the user data privacy is well protected because local training data are not shared, which thus differs FL from conventional approaches in data acquisition, storage, and training.
	
	There are several reasons why FL is attracting plenty of interests. Firstly, modern smart UEs are now able to handle heavy computing tasks of intelligent applications as they are armed with high-performance central processing units (CPUs), graphics processing units (GPUs) and integrated AI chips called neural processing units (e.g., Snapdragon 845, Kirin 980 CPU and Apple A12 Bionic CPU \cite{october22WhatDoesAI}). Being equipped with the latest computing resources at the edge, the model training can be updated locally leading to the reduction in the time to upload raw data to the data center. Secondly, the increase in storage capacity, as well as the plethora of sensors (e.g., cameras, microphones, GPS) in UEs enables them to collect a wealth amount of data and store it locally. This facilitates unprecedented large-scale flexible data collection and model training. With recent advances in edge computing, FL can be more easily implemented in reality. For example, a crowd of smart devices can proactively sense and collect data during the day hours, then jointly feedback and update the global model during the night hours, to improve the efficiency and accuracy for next-day usage. We envision that such this approach will boost a new generation of smart services, such as smart transportation, smart shopping, and smart hospital.
	
	Despite its promising benefits, FL comes with new challenges to tackle. On one hand, the number of UEs in FL can be large and the data generated by UEs have diverse distributions \cite{mcmahanCommunicationEfficientLearningDeep2017}. Designing efficient algorithms to handle statistical heterogeneity with convergence guarantee is thus a priority question. Recently, several studies \cite{mcmahanCommunicationEfficientLearningDeep2017, konecnyFederatedOptimizationDistributed2016, wangAdaptiveFederatedLearning2019} have used de facto optimization algorithms such as Gradient Descent (GD), Stochastic Gradient Descent (SGD) to enable devices' local updates in FL. One of the most well-known methods named FedAvg \cite{mcmahanCommunicationEfficientLearningDeep2017} which uses average SGD updates was experimentally shown to perform well in heterogeneous UE data settings. However, this work lacks theoretical convergence analysis. By leveraging edge computing to enable FL,  \cite{wangAdaptiveFederatedLearning2019} proposed algorithms for heterogeneous FL networks by using GD with bounded gradient divergence assumption to facilitate the convergence analysis. In another direction, the idea of allowing UEs to solve local problems in FL with arbitrary optimization algorithm to obtain a local accuracy (or inexactness level) has attracted a number of researchers \cite{smithCoCoAGeneralFramework2018, liFederatedOptimizationHeterogeneous2019}. While \cite{smithCoCoAGeneralFramework2018} uses primal-dual analysis to prove the algorithm convergence under any distribution of data, the authors of \cite{liFederatedOptimizationHeterogeneous2019} propose adding proximal terms to local functions and use primal analysis for convergence proof with a local dissimilarity assumption, a  similar idea of bounding the gradient divergence between local and global loss functions. 
	
	While all of the above FL algorithms' complexities are measured in terms of the number of local and global update rounds (or iterations), the wall clock time of FL when deployed in a wireless environment mainly depends on the number of UEs and their diverse characteristics, since UEs may have different hardware, energy budget, and wireless connection status. Specifically, the total wall-clock training time of FL includes not only the UE computation time (which depend on UEs' CPU types and local data sizes) but also the communication time of all UEs (which depends on UEs' channel gains, transmission power, and local data sizes). Thus, to minimize the wall-clock training time of FL, a careful resource allocation problem for FL over wireless networks needs to consider not only the FL parameters such as accuracy level for computation-communication trade-off, but also allocating the UEs' resources such as power and CPU cycles with respect to wireless conditions. From the motivations above, our contributions are summarized as follows:
	
	\begin{itemize}
		\item We propose a new FL algorithm with only assumption of strongly convex and smooth loss functions, named \FL. The crux of \FL is a new local surrogate function, which is designed for each UE to solve its local problem approximately up to a local accuracy level $\theta$, and is characterized by a hyper-learning rate $\eta$. 
		Using primal convergence analysis, we show the linear convergence rate of \FL by controlling $\eta$ and $\theta$, which also provides the trade-off between the number of local computation and global communication rounds. We then employ \FL, using both strongly convex and non-convex loss functions,  on PyTorch to verify its performance with several federated datasets. The experimental results show that  \FL  outperforms the vanilla FedAvg \cite{mcmahanCommunicationEfficientLearningDeep2017} in terms of training loss, convergence rate and test accuracy. 
		
		\item We propose a resource allocation problem for \FL over wireless networks to  capture the trade-off between the wall clock training time of \FL and UE energy consumption by using the Pareto efficiency model. To handle the non-convexity of this problem, we exploit its special structure to decompose it into three  sub-problems. The first two sub-problems relate to UE resource allocation over wireless networks, which are transformed to be convex and solved separately; then their solutions are used to obtain the solution to the third sub-problem, which gives the optimal $\eta$ and $\theta$ of \FL. We derive their closed-form solutions, and characterize the impact of the Pareto-efficient controlling knob to the optimal: (i) computation and communication training time, (ii) UE resource allocation, and (iii) hyper-learning rate and local accuracy. We also provide extensive numerical results to examine the impact of UE heterogeneity and Pareto curve of UE energy cost and wall clock training time.
	\end{itemize}
	
	The rest of this paper is organized as follows. Section~\ref{S:RelatedWork} discusses related works. Section~\ref{S:Model} contains system model. Sections~\ref{S:Alogrithm} and \ref{S:ResourceAll} provide the proposed FL algorithm's analysis and resource allocation over wireless networks, respectively. Experimental performance of \FL and numerical results of the resource allocation problem are provided in Section~\ref{S:Exp} and Section~\ref{S:Sim}, respectively. Section~\ref{S:Conclusion} concludes our work. 
	
	\section{Related Works}\label{S:RelatedWork}
	
	
	Due to Big Data applications and complex models such as Deep Learning, training machine learning models needs to be distributed over multiple machines, giving rise to researches on decentralized machine learning \cite{maDistributedOptimizationArbitrary2017, shamirCommunicationefficientDistributedOptimization2014,wangCooperativeSGDUnified2018a,stichLocalSGDConverges2018,zhouConvergencePropertiesKstep2018}. However, most of the algorithms in these works are designed for machines having balanced and/or independent and identically distributed (i.i.d.) data. Realizing the lack of studies in dealing with unbalanced and heterogeneous data distribution, an increasing number of researchers place interest in studying FL, a state-of-the-art distributed machine learning technique \cite{mcmahanCommunicationEfficientLearningDeep2017, konecnyFederatedLearningStrategies2016, wangAdaptiveFederatedLearning2019, smithFederatedMultitaskLearning2017,dinhFederatedLearningProximal2020,liFederatedOptimizationHeterogeneous2019}. 
	This technique takes advantage of the involvement of a large number of devices where data are generated locally, which makes them statistically heterogeneous in nature. As a result, designing algorithms with global model's convergence guarantee becomes challenging. There are two main approaches to overcome this problem.    
	
	The first approach is based on de facto algorithm SGD with a fixed number of local iterations on each device \cite{mcmahanCommunicationEfficientLearningDeep2017}. Despite its feasibility, these studies still have limitations as lacking the convergence analysis. The work in \cite{wangAdaptiveFederatedLearning2019}, on the other hand, used GD and additional assumptions on Lipschitz local functions and bounded gradient divergence to prove the algorithm convergence. 
	
	Another useful approach to tackling the heterogeneity challenge is to allow UEs to solve their primal problems approximately up to a local accuracy threshold \cite{smithFederatedMultitaskLearning2017,dinhFederatedLearningProximal2020,liFederatedOptimizationHeterogeneous2019}. Their works show that the main benefit of this approximation approach is that it allows flexibility in the compromise between the number of rounds run on the local model update and the communication to the server for the global model update. \cite{dinhFederatedLearningProximal2020} exploits proximal stochastic variance reduced gradient methods for both convex and non-convex FL. While the authors of \cite{liFederatedOptimizationHeterogeneous2019} use primal convergence analysis with bounded gradient divergence assumption and show that their algorithm can apply to non-convex FL setting, \cite{smithFederatedMultitaskLearning2017} uses primal-dual convergence analysis, which is only applicable to FL with convex problems.
	
	From a different perspective, many researchers have recently focused on the efficient  communications between UEs and edge servers in FL-supported networks \cite{wangAdaptiveFederatedLearning2019,amiriOvertheAirMachineLearning2019,tangCommunicationCompressionDecentralized,vuCellFreeMassiveMIMO2020,tranFederatedLearningWireless2019,khanFederatedLearningEdge2020a,pandeyCrowdsourcingFrameworkOnDevice2020a,chenJointLearningCommunications2020}. The work \cite{wangAdaptiveFederatedLearning2019} proposes algorithms for FL in the context of edge networks with resource constraints. While there are several works \cite{wangATOMOCommunicationefficientLearning2018,zhangZipMLTrainingLinear2017a} that study minimizing communicated messages for each global iteration update by applying sparsification and quantization, it is still a challenge to utilize them in FL networks. For example, \cite{amiriOvertheAirMachineLearning2019} uses the gradient quantization, gradient sparsification, and error accumulation to compress gradient message under the wireless multiple-access channel with the assumption of noiseless communication. The work \cite{tangCommunicationCompressionDecentralized} studies a similar quantization technique to explore convergence guarantee with low-precision training. \cite{chenJointLearningCommunications2020} considers joint learning with a subset of users and wireless factors such as packet errors and the availability of wireless resources while \cite{vuCellFreeMassiveMIMO2020} focuses on using cell-free massive MIMO to support FL. \cite{khanFederatedLearningEdge2020a,pandeyCrowdsourcingFrameworkOnDevice2020a} apply a Stackelberg game to motivate the participation of the clients during the aggregation. Contrary to most of these works which make use of existing, standard FL algorithms, our work proposes a new one. 
	Nevertheless, these works lack studies on unbalanced and heterogeneous data among UEs. We study how the computation and communication characteristics of UEs can affect their energy consumption, training time, convergence and accuracy level of FL, considering heterogeneous UEs in terms of data size, channel gain and computational and transmission power capabilities.
	
	\section{System Model} \label{S:Model}
	We consider a wireless multi-user system which consists of one edge server and a set $\Cal{N}$ of $N$ UEs. Each participating UE $n$ stores a local dataset  $\mathcal{D}_n$, with its size denoted by $D_n$. Then, we can define the total data size by $D= \sum_{n=1}^{N} D_{n}$. In an example of the supervised learning setting, at UE $n$, $\mathcal{D}_n$ defines the collection of data samples given as a set of  input-output pairs $\{x_i, y_i\}_{i=1}^{D_n}$, where $x_i \in \mathbb{R}^d$ is an input sample vector with $d$ features, and $y_i \in \mathbb{R}$ is the labeled output value for the sample $x_i$. The data can be generated through the usage of UE, for example, via interactions with mobile apps. 
	
	In a typical learning problem, for a sample data $\{x_i,y_i\}$ with input $x_i$ (e.g., the response time of various apps inside the UE), the task is to find the \emph{model parameter}  $w\in \mathbb{R}^d$ that characterizes the output $y_i$ (e.g., label of edge server load, such as high or low, in next hours) with the loss function $f_i (w)$.  The loss function on the data set of UE $n$ is defined as
	\begin{align*}
	F_n (w) \defeq \frac{1}{D_n} \SumNoLim{i \in \Cal{D}_n}{} f_i ( w). 
	\end{align*}
	Then, the learning model is the minimizer of the following global loss function minimization problem
	\begin{align}
	\min_{w \in \mathbb{R}^d} F(w) \defeq \SumNoLim{n=1}{N} p_n F_n (w), \label{E:Global_Loss}
	\end{align}
	where $p_n \defeq \frac{D_n}{D}, \forall n$. 
	
	\begin{assumption} \label{Assumption}
		$F_n(\cdot)$ is  $L$-smooth and $\beta$-strongly convex, $\forall n$, respectively, as follows, $\forall w, w' \in \mathbb{R}^d$:
		\begin{align*}
		F_n(w)    &\leq F_n(w')  \!+\!  \bigl\langle \nabla F_n (w'), w - w' \bigr\rangle\! + \! \frac{L}{2} \norm{w - w'}^2 \\
		F_n(w)  &\geq F_n(w')  \!+\! \bigl\langle \nabla F_n (w'), w - w' \bigr\rangle  \!+\! \frac{\beta }{2} \norm{w - w'}^2.  
		\end{align*}
	\end{assumption}
	Throughout this paper, $\langle w, w'\rangle$ denotes the inner product of vectors $w$ and $w'$ and $\norm{\cdot}$ is Euclidean norm. We note that strong convexity and smoothness in Assumption~\ref{Assumption}, also used in \cite{wangAdaptiveFederatedLearning2019}, can be found in a wide range of applications such as $l_2$-regularized linear regression model with $f_i (w) = \frac{1}{2} (\innProd{x_i}{w} - y_i)^2 + \frac{\beta}{2} \norm{w}^2, y_i \in \DD{R}$, and $l_2$-regularized logistic regression with  $f_i (w) = \log \bigP{ 1 + \exp ( - y_i \innProd{x_i}{w}) } + \frac{\beta}{2} \norm{w}^2, y_i \in \{-1, 1\}$. We also denote $ \rho  \defeq \frac{L}{\beta}$ the condition number of $F_n(\cdot)$'s Hessian matrix. 
	
	\section{Federated Learning Algorithm Design}\label{S:Alogrithm}
	
	In this section, we propose a FL algorithm, named \FL, as presented in Algorithm~\ref{Alg0}. To solve problem \eqref{E:Global_Loss}, \FL uses an iterative approach that requires $K_g$ \emph{global rounds}  for global model updates. In each global round, there are interactions between the UEs and edge server as follows.  
	
	\textbf{UEs update local models:} In order to obtain the local model $w_n^{t}$ at a global round $t$, each UE  $n$ first receives the feedback information $w^{t-1}$ and $\nab$ (which will be defined later in \eqref{E:Agg1} and \eqref{E:Agg2}, respectively) from the server, and then  minimize its following surrogate function (line \ref{line:cp})
	\begin{align}			
	\min_{w \in \mathbb{R}^d}  J_n^t (w) \defeq F_{n} (w) +  \bigl\langle  \eta \nab - \nabla F_n (w^{t-1} ), w \bigr\rangle.   \label{E:Computation}
	\end{align}
	One of the key ideas of \FL is  UEs can solve \eqref{E:Computation} approximately to obtain  an approximation solution $w_n^{t}$ satisfying
	\begin{align}			
	\norm{\nabla J_{n}^t (w_n^{t}) } \leq \theta \, \norm{\nabla J_{n}^t (w^{t-1}) },  \forall n, \label{E:theta_approximation}
	\end{align}
	which is parametrized by a local accuracy $\theta \in (0,1)$ that is common to all UEs. This local accuracy concept resembles the approximate factors in \cite{smithCoCoAGeneralFramework2018, reddiAIDEFastCommunication2016}. Here $\theta = 0$ means the local problem \eqref{E:Computation} is required to be solved optimally, and $\theta = 1$ means no progress for local problem, e.g., by setting $w_n^t = w^{t-1}$. The surrogate function $J_n^t(.)$ \eqref{E:Computation} is motivated from the scheme Distributed Approximate NEwton (DANE) proposed in \cite{shamirCommunicationefficientDistributedOptimization2014}. However,  DANE requires (i) the global gradient $\nabla F(w^{t-1})$ (which is not available at UEs or server in FL context),  (ii) additional proximal terms  (i.e., $\frac{\mu}{2} \norm{w - w^{t-1}}^2$), and (iii) solving local problem \eqref{E:Computation} exactly (i.e., $\theta = 0$). On the other hand, \FL uses (i) the \emph{global gradient estimate} $\nab$, which can be measured by the server from UE's information, instead of exact but unrealistic $\nabla F(w^{t-1})$, (ii) avoids using proximal terms to limit additional controlling parameter (i.e., $\mu$), and (iii) flexibly solves local problem approximately by controlling $\theta$. Furthermore, we have $\nabla J_n^t (w) = \nabla F_{n} (w) +   \eta \nab - \nabla F_n (w^{t-1} )$, which includes both local and global gradient estimate  weighted by a controllable parameter $\eta$. We will see later how $\eta$ affects to the convergence of \FL.  {Compared to the vanilla FedAvg, \FL requires more information (UEs sending not only $w_n^t$ but also $\nabla F_n (w_n^{t})$) to obtain the benefits of  a) theoretical linear convergence and b) experimental faster convergence, which will be shown in later sections. And we will also show that with Assumption~\ref{Assumption}, the theoretical analysis of \FL does not require the gradient divergence bound assumption, which is typically required in non-strongly convex cases as in \cite[Definition 1]{wangAdaptiveFederatedLearning2019}, \cite[Assumption 1]{liFederatedOptimizationHeterogeneous2019}}. 
	
	{\small 
		\begin{algorithm} [t]
			\caption{\FL} \label{Alg0}
			\begin{algorithmic}[1]
				\State \tbf{Input:} $w^{0}$, $\theta \in [0,1]$, $\eta>0$. 
				\For {$t =  1 \text{\,to\,}  K_g$}
				\State \multilines{\textbf{Computation:} Each UE $n$ receives  $w^{t-1}$ and $\nab$  from the server, and solves \eqref{E:Computation}	in $K_l$ rounds to achieve $\theta$-approximation solution $w_n^{t}$ satisfying \eqref{E:theta_approximation}.} \label{line:cp}
				\State \multilines{\textbf{Communication:} UE $n$ transmit $w_n^{t}$ and $\nabla F_n (w_n^{t} )$, $\forall n$, to the edge server.}  \label{line:co}
				\State \multilines{\textbf{Aggregation and Feedbacks:} The edge server updates the global model $w^{t}$ and $\nabla \bar{F}^{t}$ as in \eqref{E:Agg1} and \eqref{E:Agg2}, respectively,
					and then fed-backs them to all UEs.} \label{line:bs}
				\EndFor
			\end{algorithmic}
		\end{algorithm}
	}
	
	\textbf{Edge server updates global model:}
	
	After receiving the local model $w_n^t$ and gradient   $\nabla F_n (w_n^t) $, $\forall n$, the edge server aggregates them as follows
	\begin{align}
	w^{t} &\defeq  \SumN p_n  w_n^{t}, \label{E:Agg1}\\ 
	\nabla \bar{F}^{t} &\defeq \SumNoLim{n=1}{N} p_n \nabla F_n (w_n^t) \label{E:Agg2}
	\end{align}
	and then broadcast $w^t$  and  $\nabla \bar{F}^{t}$ to all UEs (line~\ref{line:bs}), which are required for participating UEs to minimize their surrogate $J_n^{t+1}$ in the next global round $t+1$. We see that the edge server does not access the local data $\Cal{D}_n, \, \forall n$, thus preserving data privacy. For an arbitrary small constant $\epsilon>0$, the  problem \eqref{E:Global_Loss}  achieves a global model convergence $w^{t}$ when its satisfies
	\begin{align}
	F(w^{t}) - F(w^*) \leq  \epsilon, \; \forall t \geq  K_g, \label{E:global}
	\end{align}
	where $w^*$ is the optimal solution to \eqref{E:Global_Loss}. 
	
	Next, we will provide the convergence analysis for \FL. We see that $J_n^t (w)$ is also $\beta$-strongly convex and $L$-smooth as $F_n(\cdot)$  because they have the same Hessian matrix. With these properties of $J_n^t (w)$, we can use GD to solve \eqref{E:Computation} as follows
	\begin{align}
	z^{k+1} = z^{k} - h_k \nabla J_{n}^t (z^{k} ), \label{E:local_gradient}
	\end{align}
	where $z_k$ is the local model update and $h_k$  is a predefined learning rate at iteration $k$,  which has been shown to generate a convergent sequence  $(z_k)_{k \geq 0}$ satisfying a linear convergence rate \cite{nesterovLecturesConvexOptimization2018} as follows
	\begin{align}
	J_{n}^t (z_k)  - J_{n}^{t} (z^*) \leq c (1 - \gamma)^k \bigP{J_{n}^t (z_0)  - J_{n}^{t} (z^*)}, \label{E:gradient_based}
	\end{align}
	where $z^*$ is the optimal solution to the local problem \eqref{E:Computation}, and $c$ and $\gamma \in (0,1)$ are constants depending on $\rho$. 
	\begin{lemma} \label{Lem:local}
		With Assumption~\ref{Assumption} and the assumed linear convergence rate \eqref{E:gradient_based} with $z_0 = w^{t-1}$, the number of local rounds $K_l$ for solving \eqref{E:Computation} to achieve a $\theta$-approximation condition \eqref{E:theta_approximation}  is
		\begin{align}
		K_l  = \frac{2}{ \gamma} \log \frac{C}{\theta },  \label{E:K_l}
		\end{align}
		where $C \defeq  c \rho $. 
	\end{lemma}
	
	\begin{theorem} \label{Th:1}
		With Assumption~\ref{Assumption}, the convergence of \FL is achieved with linear rate 
		\begin{align}
		F(w^{t}) - F(w^*) \leq (1-\Theta)^t (F(w^{(0)}) - F(w^*)), \label{E:global_convergence}
		\end{align}
		when $ \Theta \in (0,1)$, and 
		\begin{align}
		\Theta & \defeq 	\frac{\eta(2(\theta-1)^2- (\theta+1)\theta(3\eta+2)\rho^2-(\theta+1)\eta\rho^2)}{2\rho\bigP{(1+\theta)^2\eta^2\rho^2 + 1}}. \label{E:Theta}
		\end{align}
	\end{theorem}
	
	
	\begin{corollary} 
		The number of global rounds for \FL to achieve the convergence satisfying \eqref{E:global} is
		\begin{align}
		K_g  =\frac{1}{\Theta} \log \frac{F(w^0) - F(w^*)}{\epsilon}.   \label{E:K_g}
		\end{align}
	\end{corollary}
	The proof of this corollary can be shown similarly to that of Lemma~\ref{Lem:local}. We have some following remarks:
	\begin{enumerate}
		\item The convergence of \FL can always be obtained by setting sufficiently small  values of both $\eta$ and $\theta \in (0, 1)$ such that $\Theta \in (0,1)$. While the denominator  of \eqref{E:Theta} is greater than 2, its numerator can be rewritten as $2 \eta (A - B)$, where $A = 2(\theta-1)^2- (\theta+1)\theta(3\eta+2)\rho^2$ and $B = (\theta+1)\eta\rho^2$. Since $\lim_{\theta \rightarrow 0} A = 2$ and  $\lim_{\theta,\eta \rightarrow 0} B = 0$, there exists small values of $\theta$ and $\eta$ such that $A-B > 0$, thus $\Theta > 0$. On the other hand, we have $\lim_{\eta \rightarrow 0} \Theta = 0$; thus, there exists a small value of  $\eta$ such that $\Theta < 1$. We note that $\Theta \in (0,1)$ is only the sufficient condition, but not the necessary condition, for the convergence of \FL. Thus, there exist possible  hyper-parameter settings such that \FL converges but $\Theta \notin (0,1)$. 
		
		\item There is a convergence trade-off between the number of local and global rounds characterized by $\theta$:  small $\theta$ makes large $K_l$, yet small $K_g$, according to  \eqref{E:K_l} and \eqref{E:K_g}, respectively. This trade-off was also observed by authors of \cite{smithCoCoAGeneralFramework2018}, though their technique (i.e., primal-dual optimization) is different from ours. 
		
		\item While $\theta$ affects to both local and global convergence,  $\eta$ only affects to the global convergence rate of \FL. If $\eta$ is  small, then $\Theta$ is also small, thus inducing large $K_g$. However, if $\eta$ is large enough, $\Theta$ may not be in $(0,1)$, which leads to the divergence of \FL. We call $\eta$  the \emph{hyper-learning rate} for the global problem \eqref{E:Global_Loss}. 
		
		\item The condition number $\rho$ also affects to the \FL convergence: if $\rho$ is large (i.e., poorly conditioned problem \eqref{E:Computation}), both $\eta$ and $\theta$ should be sufficiently small in order for $\Theta \in (0,1)$ (i.e., slow convergence rate.) This observation is well-aligned to traditional optimization convergence analysis  \cite[Chapter 9]{boydConvexOptimization2004}. 
		
		\item In this work, the theory of \FL is applicable to (i) full data passing using GD, (ii) the participation of all UEs, and (iii) strongly convex and smooth loss functions. However, using mini-batch is a common practice in machine learning to reduce the computation load at the UEs. On the other hand,  choosing a subset of participating UEs in each global iteration is a practical approach to reduce the straggler effect, in which the run-time of each iteration  is limited by the ``slowest'' UEs (the straggler) because  heterogeneous UEs compute and communicate at different speeds. Finally, non-convex loss functions capture several essential machine learning tasks using neural networks.  In Section~\ref{S:Sim}, we will experimentally show that  \FL works well with (i) mini-batch, (ii) subset of UEs samplings, and (iii) non-convex loss functions. 

	\end{enumerate}
	
	The time complexity of \FL  is represented by  $K_g$ communication rounds and computation complexity is $K_g K_l$  computation rounds. 
	When implementing \FL over wireless networks, the wall clock time of each communication round can be significantly larger than that of computation if the number of UEs increases, due to multi-user contention for wireless medium. In the next section, we will study the UE resource allocation to enable \FL over wireless networks. 
	
	\section{\FL over Wireless Networks}\label{S:ResourceAll}
	
	
	In this section, we first present the system model and problem formulation of \FL over a time-sharing wireless environment. We then decompose this problem into three sub-problems, derive their closed-form solutions, reveal the hindsights, and provide numerical support. 
	\vspace{-0.2cm}
	\subsection{System Model}
	At first, we consider synchronous communication which requires all UEs to finish solving their local problems before entering the communication phase. During the communication phase, the model's updates are transferred to the edge server by using a wireless medium sharing scheme. In the communication phase, each global round consists of computation and communication time which includes uplink and downlink ones. In this work, however, we do not consider the downlink communication time as it is negligible compared to the uplink one. The reason is  that the downlink has larger bandwidth than the uplink and the edge server power is much higher than UE's transmission power. Besides, the computation time only depends on the number of local rounds, and thus $\theta$, according to  \eqref{E:K_l}. Denoting the  time of one local round by \Tcomp, i.e., the time to computing one local round \eqref{E:gradient_based}, then the computation time in one global round is $K_l \, \Tcomp$. Denoting the communication time in one global round by \Tcomm, the wall clock time of one global round of \FL is defined as
	\begin{align*}
	\Titer \defeq   \Tcomm +  K_l \, \Tcomp. 
	\end{align*}
	
	\subsubsection{Computation Model}
	We denote the number of CPU cycles for UE $n$ to execute one sample of data  by $c_n$, which can be measured offline \cite{miettinenEnergyEfficiencyMobile2010} and is known a priori. Since all samples $\{x_i, y_i\}_{i \in \Cal{D}_n}$ have the same size (i.e., number of bits),  the number of CPU cycles required for UE $n$ to run one local round  is $c_n D_n$. Denote the CPU-cycle frequency of the UE $n$ by $f_n$. 
	Then the CPU energy consumption  of UE $n$ for one local round of computation can be expressed as follows  \cite{burdProcessorDesignPortable1996}
	\begin{align}
	\Ecomp  = \SumNoLim{i=1}{c_n D_n} \frac{\alpha_n }{2} f_n^2 = \frac{\alpha_n }{2} c_n D_n f_n^2, \label{E:Ecomp}
	\end{align}
	where $\alpha_n/2$ is the effective capacitance coefficient of UE $n$'s computing chipset. Furthermore, the computation time per local round of the UE $n$ is $\frac{c_n D_n}{f_n}, \, \forall n.$ We denote the vector of $f_n$ by $f\in \DD{R}^n.$ 
	
	\subsubsection{Communication Model}
	In \FL, regarding to the communication phase of UEs, we consider a time-sharing multi-access protocol (similar to TDMA) for UEs. We note that this time-sharing model is not restrictive because other schemes, such as OFDMA, can also be applied to \FL. The achievable transmission rate (nats/s) of UE $n$ is defined as follows:
	\begin{align}
	r_n = B \ln \bigP{ 1 + \frac{\bar{h}_n p_n}{N_0}}, \label{E:rate1}
	\end{align}
	where $B$ is the bandwidth, $N_0$ is the background noise, $p_n$ is the transmission power, and $\bar{h}_n$ is the average channel gain of the UE $n$ during the training time of \FL.
	Denote the fraction of communication time  allocated to UE $n$ by $\tau_n$, and the data size (in nats) of  $w_n$ and $\nabla F_n(w_n)$  by $s_n$.  Because the dimension of vectors $w_n$ and $\nabla F_n(w_n)$ is fixed, we assume that their sizes are constant throughout the \FL learning. Then the transmission rate of each UE $n$ is 
	\begin{align}
	r_n = s_n/\tau_n, \label{E:rate2}
	\end{align}
	which is shown to be the most energy-efficient transmission policy  \cite{prabhakarEnergyefficientTransmissionWireless2001}. Thus, to transmit $s_n$ within a time duration $\tau_n$, the UE $n$'s energy consumption is
	\begin{align}
	\Ecomm =   \tau_n \, p_n(s_n/\tau_n), \label{E:Ecomm}
	\end{align}
	where the power function is 
	\begin{align}
	p_n(s_n/\tau_n) \defeq \frac{N_0}{\bar{h}_n}  \BigP{e^{\frac{s_n/\tau_n}{B}} - 1} \label{E:p_func_of_tau}
	\end{align}
	according to \eqref{E:rate1} and \eqref{E:rate2}. We denote the vector of $\tau_n$ by $\tau \in \DD{R}^n$. 
	
	Define the total energy consumption of all UEs for each global round by $\Eiter$, which is expressed as follows:
	\begin{align}
	\Eiter  \defeq \SumNoLim{n=1}{N} \Ecomm + K_l \, \Ecomp. \nonumber
	\end{align}
	
	\subsection{Problem formulation}
	We consider an optimization problem, abusing the same name \FL, as follows
	\begin{align}	
	& \underset{f, \tau, \theta, \eta, \Tcomm, \Tcomp}{\text{minimize}} && K_g \bigP{\Eiter + \kappa \, \Titer } \nonumber\\
	& \text{subject to}
	&& \SumNoLim{n=1}{N} \tau_n \leq \Tcomm, \; \label{E:constr_comm} \\
	&&& \max_n \frac{c_n D_n}{f_n} = \Tcomp, \;  \label{E:constr_comp}\\
	& && \fMin \leq f_n \leq \fMax, \; \forall n \in \Cal{N}, \label{E:freq}\\
	&&& \pMin \leq p_n(s_n/\tau_n)  \leq \pMax, \; \forall n \in \Cal{N}, \label{E:power}\\
	&&&   0 \leq \theta \leq  1. 
	\end{align}	
	
	Minimize both UEs' energy consumption and the FL time are conflicting. For example, the UEs can save the energy by setting the lowest frequency level all the time, but this will certainly increase the training time. Therefore, to strike the balance between energy cost and training time, the weight $\kappa$ (Joules/second), used in the objective as an amount of additional energy cost that \FL is willing to bear for one unit of training time to be reduced,  captures the Pareto-optimal tradeoff between the UEs' energy cost and the FL time. For example, when most of the UEs are plugged in, then UE energy is not the main concern, thus $\kappa$ can be large. According to optimization theory, $1/\kappa$ also plays the role of a Lagrange multiplier for a ``hard constraint'' on UE energy \cite{boydConvexOptimization2004}. 
	
	While constraint \eqref{E:constr_comm} captures the time-sharing uplink transmission of UEs, constraint \eqref{E:constr_comp}  defines that  the computing time in one local round is determined by the ``bottleneck'' UE (e.g., with large data size and low CPU frequency). The feasible regions of CPU-frequency and transmit power of UEs are imposed by constraints  \eqref{E:freq} and \eqref{E:power}, respectively.  We note that  \eqref{E:freq} and \eqref{E:power} also capture the heterogeneity of UEs with different types of CPU and transmit chipsets. The last constraint restricts the feasible range of the local accuracy. 
	
	\vspace{-0.5cm}
	\subsection{Solutions to \Opt} \label{S:Sols}
	
	We see that  \Opt is non-convex due to the constraint \eqref{E:constr_comp} and several products of two functions in the objective function. However, in this section we will characterize \Opt's solution by decomposing it into multiple simpler sub-problems. 
	
	We consider the first case when $\theta$ and $\eta$ are fixed, then \Opt can be decomposed into two sub-problems as follows:
	\begin{align}	
	\SubOne\! \!: \quad
	& \underset{f,  \Tcomp}{\text{minimize}} &&  \SumNoLim{n=1}{N} \Ecomp  + \kappa \Tcomp \nonumber \\
	& \text{subject to}
	&& \frac{c_n D_n}{f_n} \leq \Tcomp, \; \forall n \in \mathcal{N}, \label{E:constr_comp'}\\
	& && \fMin \leq f_n \leq \fMax, \; \forall n \in \Cal{N}.  \nonumber 
	\end{align}	
	\vspace{-0.7cm}
	\begin{align}	
	\SubTwo\!: \quad 
	& \underset{ \tau, \Tcomm}{\text{min.}} &&  \SumNoLim{n=1}{N} \Ecomm +  \kappa  \Tcomm  \nonumber \\
	& \text{s.t.}
	&& \SumNoLim{n=1}{N} \tau_n \leq \Tcomm, \; \label{E:constr_comm''} \\
	&&&\pMin \leq p_n(s_n/\tau_n)  \leq \pMax, \, \forall n. \label{E:constr_p} 
	\end{align}	
	
	While \SubOne is a  CPU-cycle control problem for the computation time and energy minimization,  \SubTwo can be considered as an uplink power control to determine the UEs' fraction of time sharing to minimize the UEs energy and communication time. We note that the constraint \eqref{E:constr_comp} of \Opt is replaced by an equivalent one \eqref{E:constr_comp'} in \SubOne. We can consider \Tcomp\ and \Tcomm\ as virtual deadlines for UEs to perform their computation and communication updates, respectively.  It can be observed that both \SubOne and \SubTwo are convex problems. 
	
	\subsubsection{\SubOne Solution}
	We first propose Algorithm~\ref{Alg1} in order to categorize UEs into one of three groups: \None\ is a group of ``bottleneck'' UEs that always run its maximum frequency;  \Ntwo\ is the group of ``strong'' UEs which can finish their tasks before  the computational virtual deadline even with the minimum frequency; and  \Nthree\ is the group of UEs having the optimal frequency inside the interior of their feasible sets.
	\begin{figure*}[!t]
		\centering
		\begin{subfigure}{0.25\textwidth} 
			\includegraphics[width=1.\linewidth]{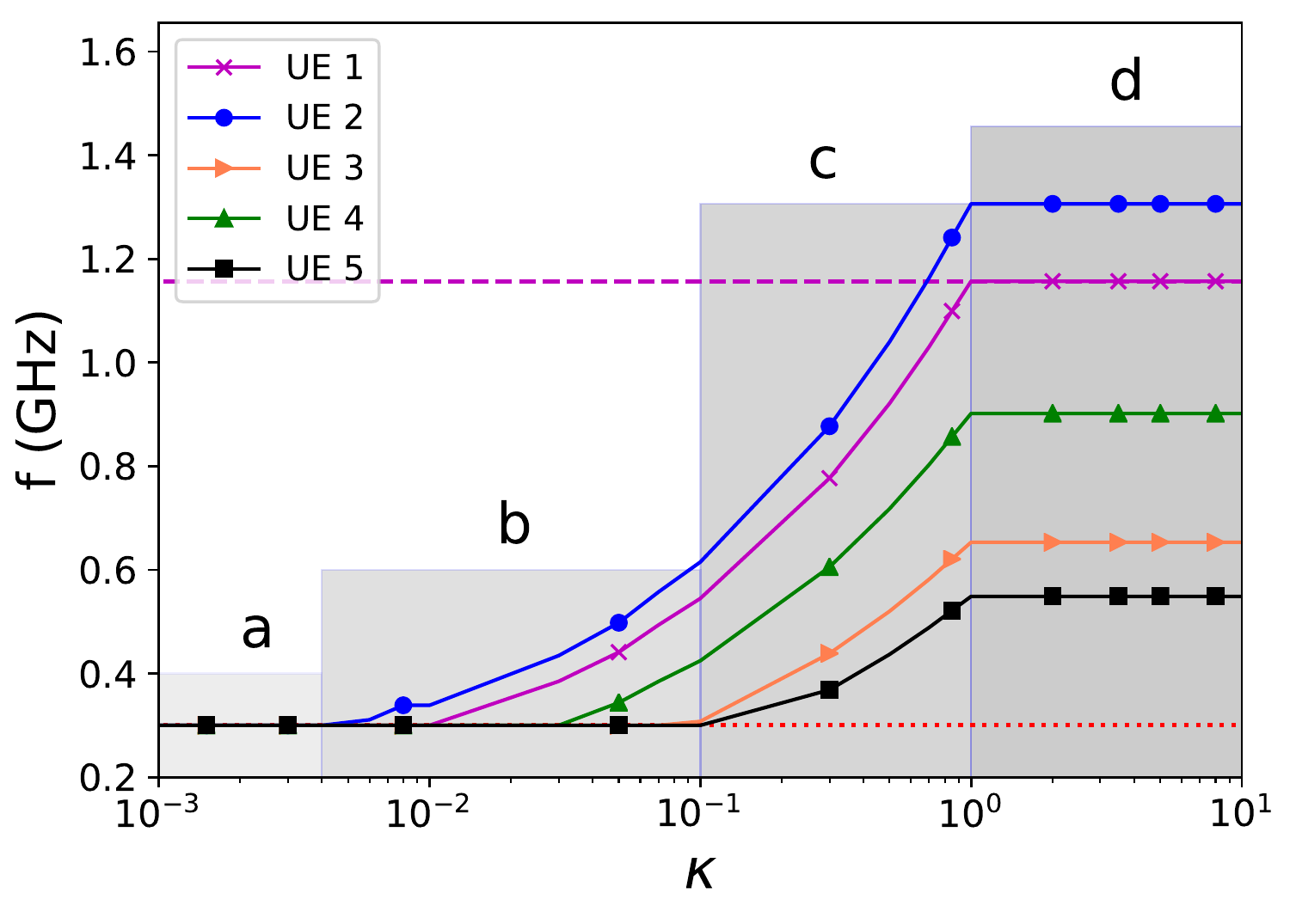}
			\caption{Optimal CPU frequency of each UE.}
			\label{F:Sub1_a}
		\end{subfigure}\hfil
		\begin{subfigure}{0.25\textwidth} 
			\includegraphics[width=1.\linewidth]{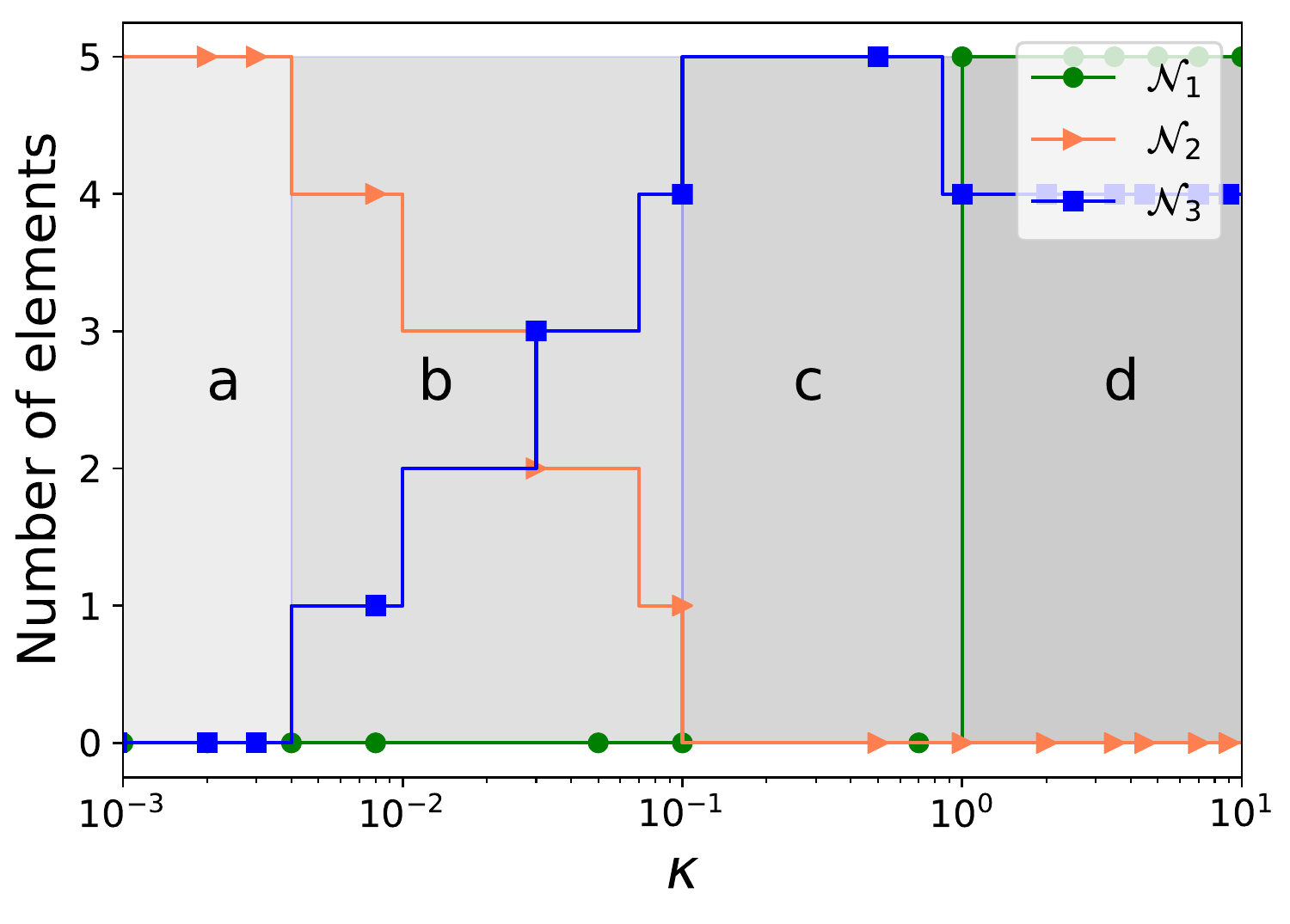}
			\caption{Three subsets outputted by Alg.~\ref{Alg1}.}
			\label{F:Sub1_b}
		\end{subfigure} \hfil 
		\begin{subfigure}{0.25\textwidth} 
			\includegraphics[width=1.\linewidth]{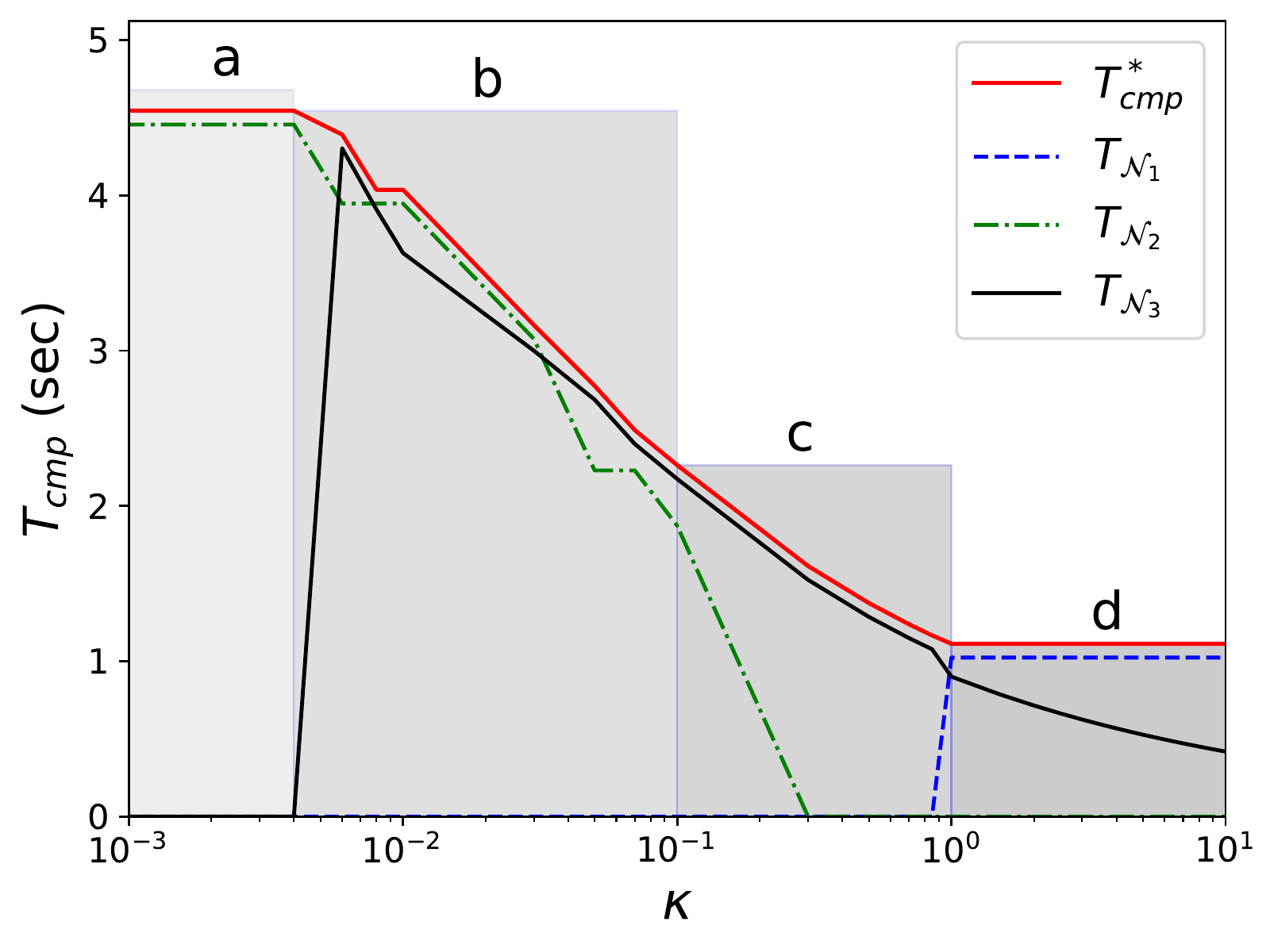}
			\caption{Optimal computation time.}
			\label{F:Sub1_c}
		\end{subfigure}
		\caption{Solution to \SubOne with five UEs. For wireless communication model, the UE channel gains follow the exponential distribution with the mean $g_0(d_0/d)^4$ where $g_0 = -40$ dB and the reference distance $d_0 = 1$ m. The distance between these devices and the wireless access point is uniformly distributed between $2$ and $50$ m. In addition, $B = 1$ MHz, $\sigma = 10^{-10}$ W, the transmission power of devices are limited  from $0.2$ to $1$ W. For UE computation model, we set the training size $D_n$ of each UE  as uniform distribution in $5 - 10$ MB, $c_n$ is uniformly distributed in $10 - 30$ cycles/bit, $f^{max}_n$  is uniformly distributed in $1.0 - 2.0$ GHz, $f^{min}_n = 0.3$ GHz. Furthermore, $\alpha = 2 \times 10^{-28}$ and the UE update size $s_n = 25,000$ nats ($\approx$4.5 KB).   } 	\label{F:Sub1}
	\end{figure*}
	
	\begin{lemma} \label{L:1}
		The optimal solution to \SubOne is as follows
		\begin{align}	
		f_n^* &= \begin{cases}  \label{E:Lemm1_a_2}
		\fMax,   &\forall {n \in \Cal{N}_1}, \\		
		\fMin,   &\forall {n \in \Cal{N}_2}, \\
		\frac{c_n D_n}{\TcompOpt},   &\forall {n \in \Cal{N}_3}, \\
		\end{cases}\\
		\TcompOpt &= \max \bigC{ T_{\Cal{N}_1}, T_{\Cal{N}_2}, T_{\Cal{N}_3} }, 	\label{E:ToptLemma1}
		\end{align}
		where $\Cal{N}_1, \Cal{N}_2, \Cal{N}_3 \subseteq \Cal{N}$ are three subsets of UEs produced by Algorithm~\ref{Alg1} and
		\begin{align}
		T_{\Cal{N}_1} &= \max_{n \in \Cal{N}}\frac{c_n D_n}{\fMax},  \nonumber\\
		T_{\Cal{N}_2} &= \max_{n \in \Cal{N}_2}\frac{c_n D_n}{\fMin},   \nonumber \\
		T_{\Cal{N}_3} &= \biggP{\frac{\SumNoLim{n \in \Cal{N}_3}{} \alpha_n (c_n D_n)^3 }{\kappa }}^{1/3}. \label{E:T_N3}
		\end{align}	
		
		%
	\end{lemma}
	{\small 
		\begin{algorithm} [t]
			\caption{Finding $\Cal{N}_1, \Cal{N}_2, \Cal{N}_3$ in  Lemma~\ref{L:1}} \label{Alg1}
			\begin{algorithmic}[1]
				\State Sort UEs such that  $\frac{c_1 D_1}{f_1^{min}} \leq  \frac{c_{2} D_{2}} {f_{2}^{min}} \ldots \leq \frac{c_{N} D_{N}} {f_{N}^{min}}$ 
				\State \textbf{Input:} $\; \Cal{N}_1 = \emptyset$, $\Cal{N}_2 = \emptyset, \, \Cal{N}_3 = \Cal{N}$, $T_{\Cal{N}_3}$ in \eqref{E:T_N3}
				\For {$i =  1 \text{\,to\,}  N$}
				\If {$\max_{n\in \Cal{N}} \frac{c_{n} D_{n }}{f_{n}^{max}} \geq T_{\Cal{N}_3}>0$ and $\None == \emptyset$} \label{Alg1:ifN1}
				\State $\Cal{N}_1  = \Cal{N}_1 \cup \bigC{m: \frac{c_{m} D_{m}}{f_{m}^{max}} =  \max_{n\in \Cal{N}} \frac{c_{n} D_{n }}{f_{n}^{max}} }$ \label{Alg1:N1add}
				\State $\Cal{N}_3  = \Cal{N}_3 \setminus \Cal{N}_1$ and update $T_{\Cal{N}_3}$ in \eqref{E:T_N3}
				\EndIf
				\If {$ \frac{c_i D_i}{f_i^{min}} \leq T_{\Cal{N}_3}$} \label{Alg1:while}
				\State	$\Cal{N}_2  = \Cal{N}_2 \cup \{i\}$ \label{Alg1:N2add}
				\State $\Cal{N}_3  = \Cal{N}_3 \setminus \{i\}$ and update $T_{\Cal{N}_3}$ in \eqref{E:T_N3} \label{Alg1:N3add}
				\EndIf
				\EndFor
			\end{algorithmic}
		\end{algorithm}
	}
	From Lemma~\ref{L:1}, first, we see that the optimal solution  depends not only on the existence of these subsets, but also on their virtual deadlines \TnOne, \TnTwo, and \TnThree, in which the longest of them will determine the optimal virtual deadline \TcompOpt. Second, from \eqref{E:Lemm1_a_2}, the optimal frequency of each UE will depend on both \TcompOpt\ and the subset it belongs to.  We note that depending on $\kappa$,  some of the three sets (not all) are possibly empty sets, and by default $T_{\Cal{N}_i} = 0$ if $\Cal{N}_i$ is an empty set, $i =1, 2, 3$. Next, by varying $\kappa$, we observe the following special cases. 
	
	\begin{corollary} \label{C:1}The optimal solution to \SubOne can be divided into four regions as follows.  
		\begin{enumerate}[a)]
			\item $\kappa \leq \min_{n \in \Cal{N} }\alpha_n {(\fMin)}^3:$\\
			\None\ and \Nthree\ are empty sets. Thus, $\Ntwo = \Cal{N}$, $\TcommOpt = \TnTwo = \max_{n \in \Cal{N}}\frac{c_n D_n}{\fMin}$, and $f_n^* = \fMin,  \forall n \in \Cal{N}$. 		
			
			\item $ \min_{n \in \Cal{N} }\alpha_n {(\fMin)}^3 < \kappa \leq \bigP{\max_{n \in \Ntwo} \frac{c_n D_n}{\fMin}}^3:$\\
			\Ntwo\ and \Nthree\ are non-empty sets, whereas \None\ is empty. Thus, $\TcompOpt = \max\bigC{\TnTwo, \TnThree}$, and $f_n^* = \max \bigC{\frac{c_n D_n}{\TcompOpt}, \fMin}, \forall {n \in \Cal{N}}$. 
			
			\item $\bigP{\max_{n \in \Ntwo} \frac{c_n D_n}{\fMin}}^3 < \kappa \leq \frac{\SumNoLim{n \in \Nthree}{} \alpha_n \bigP{c_n D_n}^3}{\bigP{\max_{n \in \Cal{N}} \frac{c_n D_n}{\fMax}}^3}$: \\
			\None\ and \Ntwo\ are empty sets. Thus $\Nthree = \Cal{N}$, $\TcompOpt = \TnThree$, and $f_n^* = \frac{c_n D_n}{\TnThree}, \forall n \in \Cal{N}$.
			
			\item $\kappa > \frac{\SumNoLim{n \in \Nthree}{} \alpha_n \bigP{c_n D_n}^3}{\bigP{\max_{n \in \Cal{N}} \frac{c_n D_n}{\fMax}}^3}$:\\
			\None\ is non-empty. Thus $\TcompOpt = \TnOne$, and 
			\begin{align}	
			f_n^* &= \begin{cases} \label{E:Corollary1}
			\fMax,   &\forall {n \in \None,} \\
			\max \bigC{\frac{c_n D_n}{\TnOne}, \fMin},   &\forall {n \in \Cal{N} \setminus \None.} 
			\end{cases}
			\end{align}	  	

		\end{enumerate}
	\end{corollary}
	
	We illustrate  Corollary~\ref{C:1} in  Fig.~\ref{F:Sub1} with four regions\footnote{All closed-form solutions are also verified by the solver IPOPT \cite{wachterImplementationInteriorpointFilter2006}.} as follows. 
	
	\begin{enumerate}[a)]
		\item Very low $\kappa$ (i.e., $ \kappa \leq 0.004$): Designed for solely energy minimization. In this region, all UE runs their CPU at the lowest cycle frequency \fMin,  thus \TcompOpt\ is  determined by the last UEs that finish their computation with their minimum frequency.  
		
		\item Low $\kappa$ (i.e., $ 0.004 \leq \kappa \leq 0.1$): Designed for prioritized energy minimization. This region contains UEs of both \Ntwo\ and \Nthree. \TcompOpt\ is  governed by which subset has higher  virtual computation deadline, which also determines the optimal CPU-cycle frequency of $\Nthree$. Other UEs with light-loaded data, if exist, can run at the most energy-saving mode \fMin\ yet still finish their task before \TcompOpt\ (i.e., \Ntwo). 
		
		\item Medium $\kappa$ (i.e., $ 0.1 \leq \kappa \leq 1$): Designed  for balancing computation time and energy minimization. All UEs belong to \Nthree\ with their optimal CPU-cycle frequency strictly inside the feasible set.  
		
		\item High $\kappa$ (i.e., $\kappa \geq 1$): Designed for prioritized computation time minimization. High value $\kappa$ can ensure the existence of \None, consisting the most ``bottleneck'' UEs (i.e., heavy-loaded data and/or low \fMax)  that runs their maximum CPU-cycle in \eqref{E:Corollary1} (top) and thus determines the optimal computation time \TcompOpt. The other ``non-bottleneck'' UEs  either (i) adjust a ``right'' CPU-cycle  to save the energy yet still maintain their computing time the same as \TcompOpt\ (i.e., \Nthree), or (ii) can finish the computation with minimum frequency before the ``bottleneck'' UEs (i.e., \Ntwo) as in \eqref{E:Corollary1} (bottom).
	\end{enumerate}

	
	\subsubsection{\SubTwo Solution}
	Before characterizing the solution to \SubTwo, from \eqref{E:p_func_of_tau} and \eqref{E:constr_p}, we first define two bounded values for $\tau_n$ as follows
	\begin{align*}
	\tauMax &= \frac{s_n}{B \ln (\bar{h}_n N_0^{-1} \pMin+ 1) },	\\
	\tauMin 	&=  \frac{s_n}{B \ln (\bar{h}_n N_0^{-1} \pMax+ 1) }, 
	\end{align*}
	which are the maximum and minimum possible fractions of \Tcomm\ that UE $n$ can achieve by transmitting with its minimum and maximum power, respectively.  We also define a new function $g_n: \DD{R} \rightarrow \DD{R}$  as
	\begin{align*}
	g_n (\kappa) = \frac{s_n/B} {1 +{W \bigP{ \frac{ \kappa N_0^{-1} \bar{h}_n - 1}{e} }}},
	\end{align*}
	where $W(\cdot)$ is the Lambert $W$-function. We can consider $g_n(\cdot)$ as an indirect ``power control'' function that helps UE $n$ control the amount of time it should transmit an amount of data $s_n$ by adjusting the power based on the weight $\kappa$. This function is strictly decreasing (thus its inverse function $g_n^{-1}(\cdot)$ exists) reflecting that when we put more priotity on minimizing the communication time (i.e., high $\kappa$), UE $n$ should raise the power to finish its transmission with less time (i.e., low $\tau_n$). 
	
	\begin{lemma} \label{L:2}
		The solution to \SubTwo \,  is as follows
		\begin{enumerate}[a)]
			\item  If $\kappa \leq  g_n^{-1}(\tauMax) $,  then
			\begin{align*}
			\tau_{n}^* &= \tauMax 
			\end{align*}
			
			\item 	 If $ g_n^{-1}(\tauMax) < \kappa < g_n^{-1}(\tauMin) $, then 
			\begin{align*}
			\tauMin <	\tau_{n}^*  &= g_n (\kappa) < \tauMax
			\end{align*}
			
			\item   If $\kappa \geq  g_n^{-1}(\tauMin)$, then
			\begin{align*}
			\tau_{n}^* &= \tauMin, 
			\end{align*}
		\end{enumerate}	
		and $ \TcommOpt = \SumNoLim{n=1}{N} \tau_{n}^*. $
	\end{lemma}
	
	\begin{figure}
		\centering
		\begin{subfigure}{0.45\linewidth} 
			\includegraphics[width=1.\linewidth]{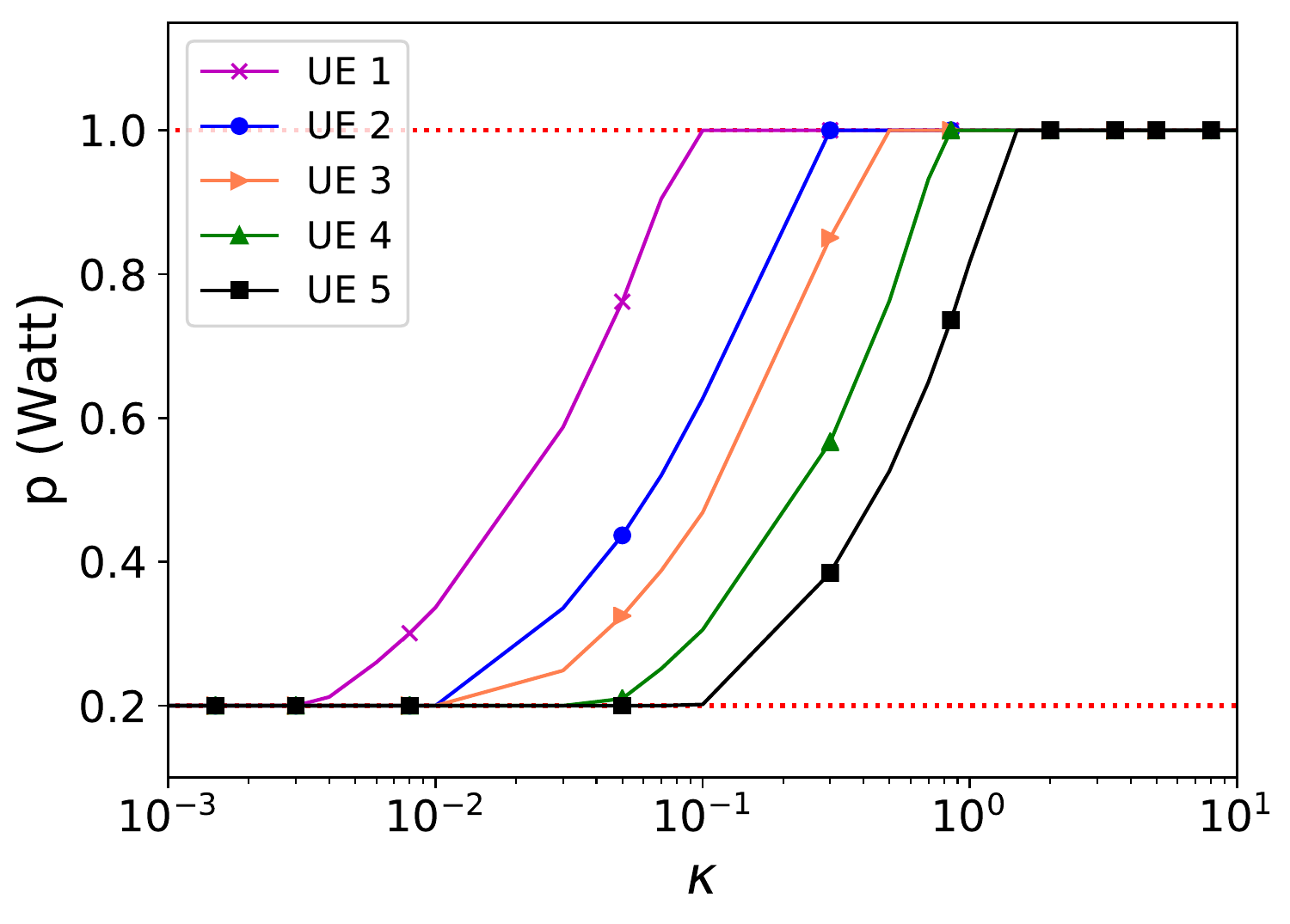}
			\caption{UEs' optimal transmission power.}
			\label{F:Sub2_p}
		\end{subfigure}
		\begin{subfigure}{0.45\linewidth} 
			\includegraphics[width=1.\linewidth]{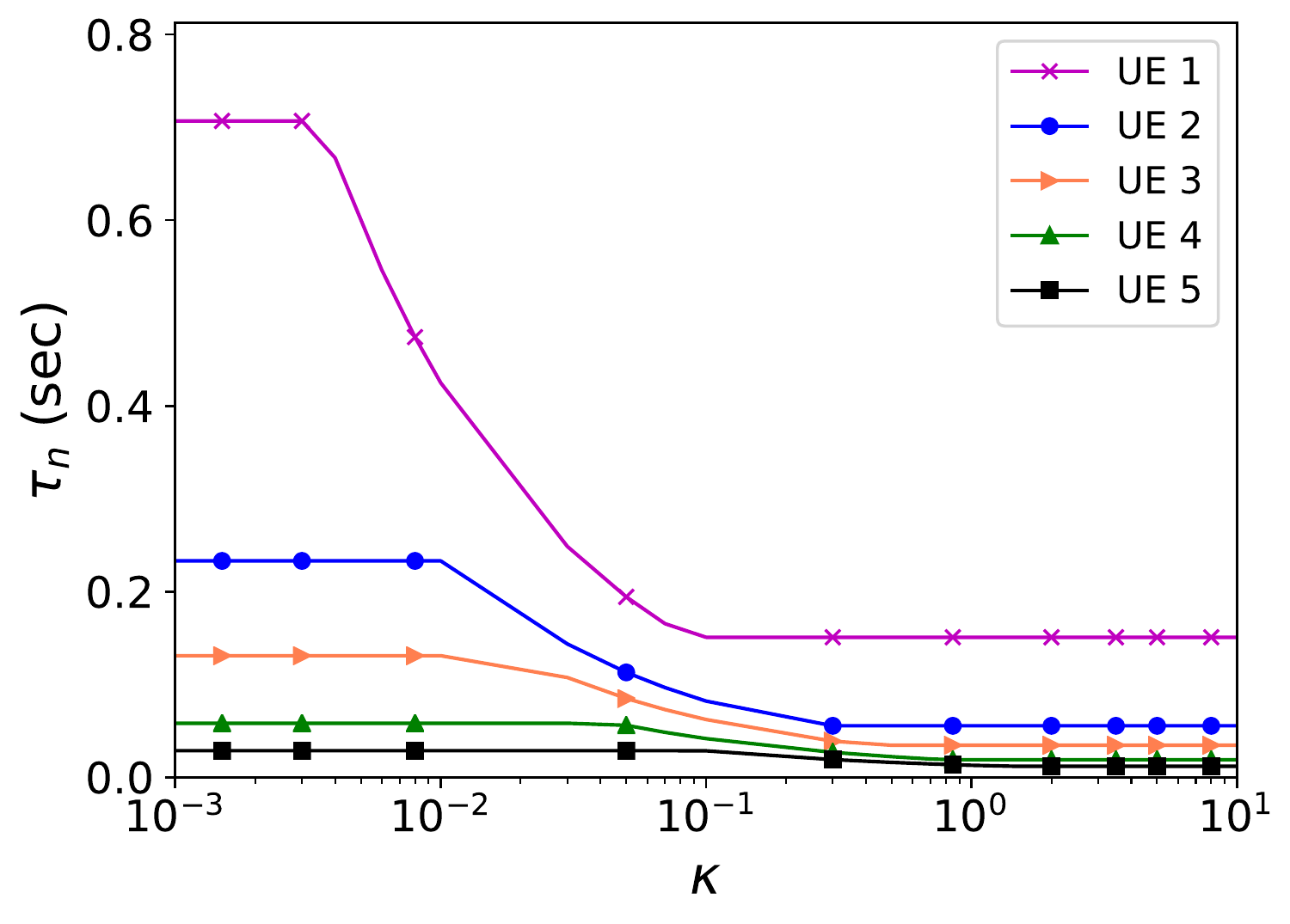}
			\caption{UEs' optimal transmission time .}
			\label{F:Sub2_tau}
		\end{subfigure}
		\caption{The solution to \SubTwo with five UEs. The numerical setting is the same as that of Fig.~\ref{F:Sub1}.}  \label{F:Sub2}
	\end{figure}

	\begin{table}[ht]	
	\centering
	\begin{tabular}{ |c|c|c|c|c|c| } 
		\hline
		$\rho$ &  $\kappa$ & $\theta^*$ & $\Theta$ & $\eta^*$  \\
		\hline
		\multirow{3}{*}{1.4/2/5} 
		& 0.1 &  .033/.015/.002 & .094/.042/.003 & .253/.177/.036  \\\cline{2-5} 
		& 1 & .035/.016/.002  & .092/.041/.003  & .253/.177/.036   \\ \cline{2-5} 
		& 10 & .035/.016/.002  & .092/.041/.003 & .253/.177/.036   \\ 
		\hline
	\end{tabular}  
	\caption{The solution to \SubThree with five UEs. The numerical setting is the same as that of Fig.~\ref{F:Sub1}.} \label{T:SUB3}
\end{table}

	This lemma can be  explained through the lens of network economics. If we interpret the \FL\ system as the buyer and UEs as sellers with the UE powers as commodities, then the inverse function $g_n^{-1}(\cdot)$ is interpreted as the price of energy that UE $n$ is willing to accept to provide power service for \FL\ to reduce the training time. There are two properties of this function: (i) the price increases with respect to UE power, and (ii) the price sensitivity depends on UEs characteristics, e.g., UEs with better channel quality can have lower price, whereas UEs with larger data size $s_n$ will have higher price. Thus, each UE $n$ will compare its energy price $g_n^{-1}(\cdot)$ with the ``offer'' price $\kappa$ by the system to decide how much power it is willing to ``sell''. Then, there are three cases corresponding to the solutions to \SubTwo. 
	
	\begin{enumerate}[a)]
		\item Low offer: If the offer price $\kappa$  is lower than the minimum price request $g_n^{-1}(\tauMax)$,  UE $n$ will sell its lowest service by transmitting with the minimum power $\pMin$. 
		
		\item Medium offer: If the offer price $\kappa$ is within the range of an acceptable price range, UE $n$ will find a power level such that the corresponding energy price will match the offer price. 
		
		\item High offer: If the offer price $\kappa$  is higher than the maximum price request $g_n^{-1}(\tauMin)$,  UE $n$ will sell its highest service by transmitting with the maximum power $\pMax$. 
	\end{enumerate}
	
	Lemma~\ref{L:2} is further illustrated in Fig.~\ref{F:Sub2}, showing how the solution to \SubTwo varies with respect to $\kappa$. It is observed from this figure that due to the UE heterogeneity of channel gain, $\kappa = 0.1$ is a medium offer to UEs 2, 3, and 4, but  a high offer to UE 1, and low offer to UE 5. 
	
	While \SubOne and \SubTwo solutions share the same threshold-based dependence, we observe their differences as follows. In \SubOne solution,  the optimal CPU-cycle frequency of UE $n$ depends on the optimal \TcompOpt, which in turn depends on the loads (i.e., $\frac{c_n D_n}{f_n}, \, \forall n$ ) of all UEs. Thus all UE load information is required for the computation phase. On the other hand, in \SubTwo solution, each UE $n$ can independently choose its optimal power by comparing its price function $g_n^{-1}(\cdot)$ with $\kappa$ so that collecting UE information is not needed.  The reason is that the synchronization of computation time in constraint \eqref{E:constr_comp'} of \SubOne requires all UE loads, whereas the  UEs' time-sharing constraint \eqref{E:constr_comm''} of \SubTwo can be decoupled by comparing with the fixed ``offer'' price $\kappa$.

	\subsubsection{\SubThree Solution}
	We observe that the solutions to \SubOne and \SubTwo have no dependence on $\theta$ so that the optimal  \TcommOpt, \TcompOpt, $f^*$, $\tau^*$, and thus the corresponding optimal energy values, denoted by \EcompOpt~ and \EcompOpt, can be determined based on $\kappa$ according to Lemmas \ref{L:1} and \ref{L:2}.  However, these solutions will affect to the third sub-problem of \FL, as will be shown  in what follows. 
	\begin{align}	
	&\SubThree \!:  \nonumber \\
	&\underset{ \theta, \eta>0}{\text{minimize}} &&  \frac{1}{\Theta} \BigP{\SumNoLim{n=1}{N} {\EcommOpt} + K_l \, {\EcompOpt} + \kappa \big({\TcommOpt} + K_l \, {\TcompOpt}\big) } \nonumber\\
	& \text{subject to} &&   0 <  \theta <  1, 0 < \Theta < 1. \ \nonumber
	\end{align}	
	
	\SubThree is unfortunately non-convex. However, since there are only two variables to optimize, we can employ numerical methods to find the optimal solution. The numerical results in Table~\ref{T:SUB3} show that the solution $\theta^*$ and $\eta^*$ to \SubThree decreases when $\rho$ increases, which makes $\Theta$ decreases, as explained by the results of Theorem~\ref{Th:1}. Also we observe that $\kappa$ as more effect to the solution to \SubThree when $\rho$ is small. 
	
	\subsubsection{\Opt Solution} Since we can obtain the stationary points of \SubThree using Successive Convex Approximation techniques such as NOVA \cite{scutariParallelDistributedMethods2017b}, then we have:
	\begin{theorem}
		The combined solutions to three sub-problems \SubOne, \SubTwo, and \SubThree are stationary points of \Opt.  
	\end{theorem}
	
	The proof of this theorem is straightforward. The idea is to use the KKT condition to find the stationary points of \Opt. Then we can decompose the KKT condition  into three independent groups of equations (i.e., no coupling variables between them), in which the first two groups matches exactly to the KKT conditions of  \SubOne and \SubTwo that can be solved by closed-form solutions as in Lemmas~\ref{L:1}, \ref{L:2}, and the last group for \SubThree is solved by numerical methods. 
	
	We then have some discussions on the combined solution to \Opt. First, we see that \SubOne and \SubTwo solutions can be characterized independently, which can be explained that  each UE often has two separate processors: one CPU for mobile applications and another baseband processor for radio control function. 
	Second, neither \SubOne nor \SubTwo depends on $\theta$ because  the communication phase in \SubTwo is clearly not affected by the local accuracy, whereas \SubTwo considers the computation cost in one local round. However, the solutions to \SubOne and \SubTwo, which can reveal how much communication cost is more expensive than computation cost, are decisive factors to determine the optimal level of local accuracy. Therefore, we can sequentially solve \SubOne and \SubTwo first, then \SubThree to achieve the solutions to \Opt. We also summarize the complexities in the following table:
		\begin{table}[H]	
			\centering
			\begin{tabular}{ |c|c|c| } 
				\hline
				SUB1 &  SUB2 & SUB3 \\
				\hline
				$O(N^2) $& $O(1) $& $O(N)$\\
				\hline
			\end{tabular}  
			\caption{Summary of the sub-problem complexity.} \label{T:SUB}
	\end{table}
	
	\vspace{-0.2cm}
	\section{Experiments} \label{S:Exp}
	\begin{figure}[t!]
		\centering
		\includegraphics[width=1\linewidth]{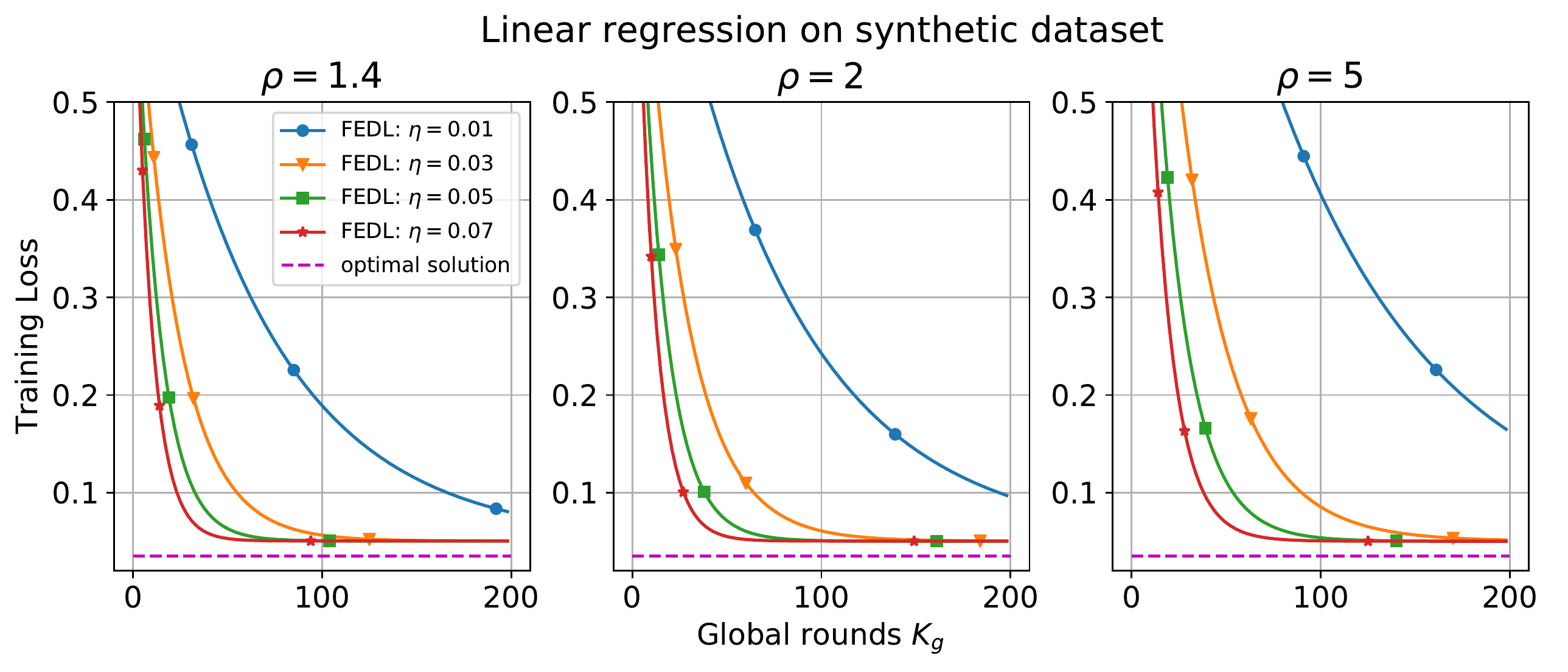}\\
		\caption{Effect of $\eta$ on the convergence of \FL. Training processes use full-batch gradient descent, full devices participation ($N = S = 100$ UEs), $K_g = 200$, and $K_l = 20$.}
		\label{F:effects_of_eta}
	\end{figure}
	\begin{figure}[t!]
		\centering
		\includegraphics[width=1\linewidth]{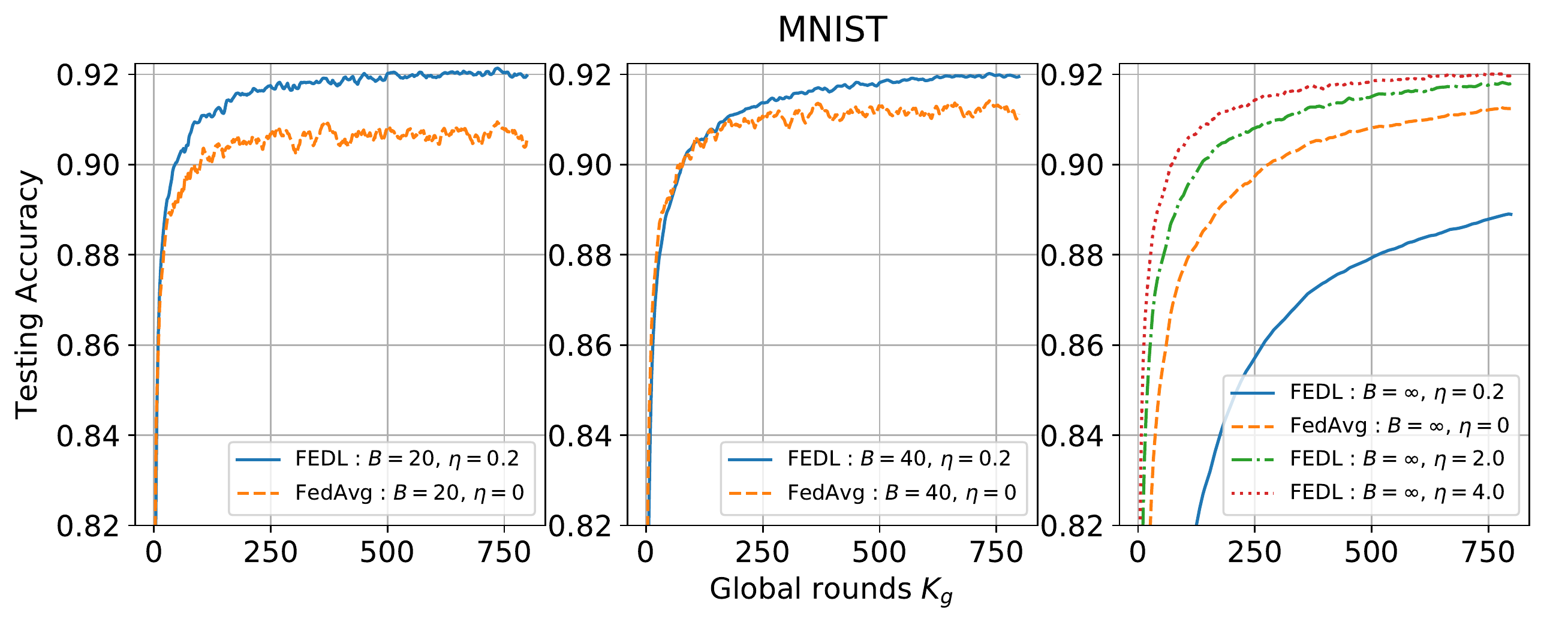}\\
		\includegraphics[width=1\linewidth]{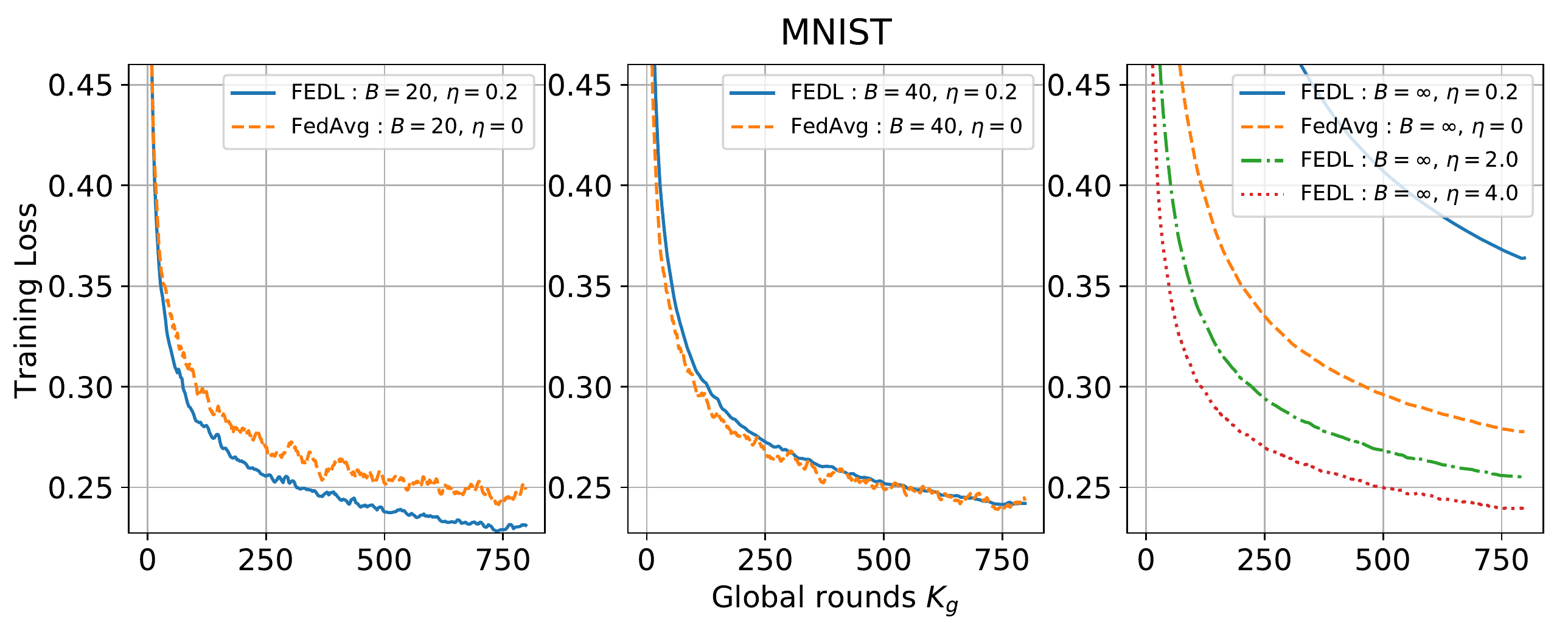}\\
		\caption{Effect of different batch-size with fixed value of $K_l = 20$, $K_g = 800$, $N = 100$, $S = 10$, ($B = \infty$ means full batch size) on \FL's performance.}
		\label{F:nist_local_batch}
	\end{figure}
	\begin{figure}[t!]
		\centering
		\includegraphics[width=1\linewidth]{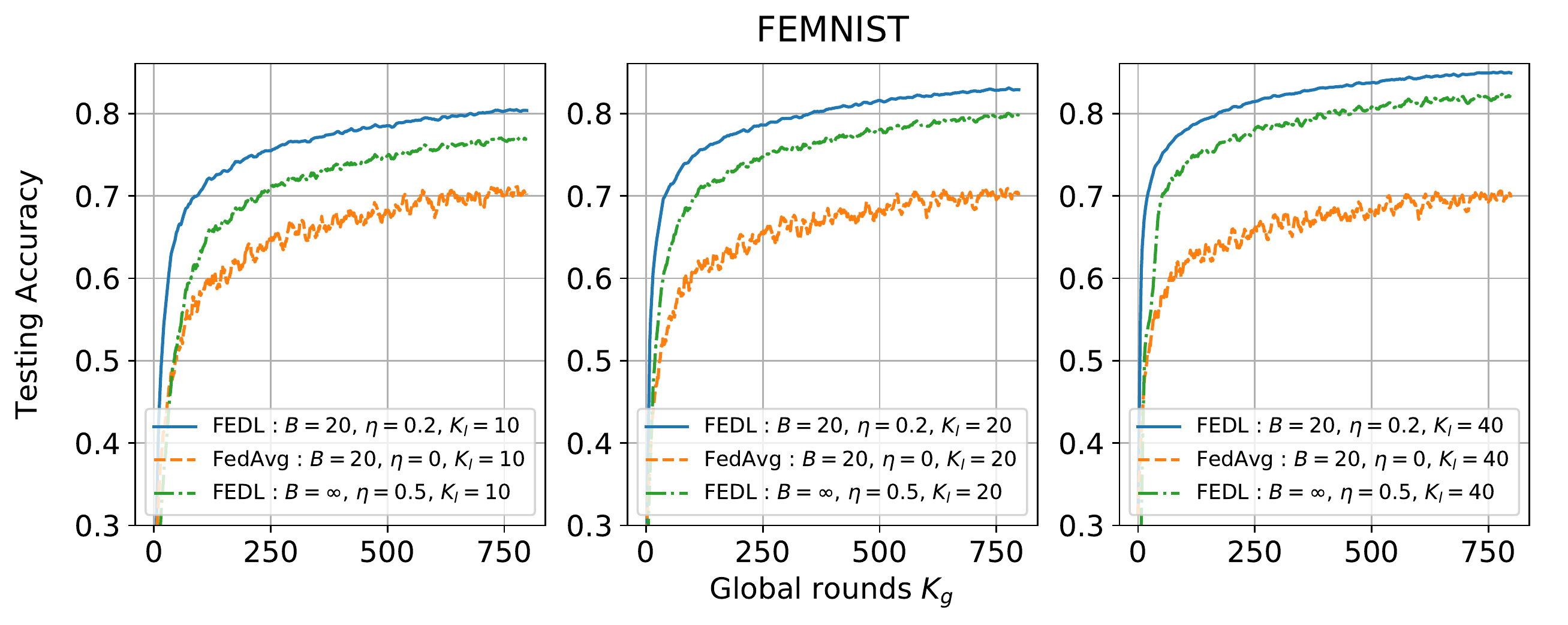}\\
		\includegraphics[width=1\linewidth]{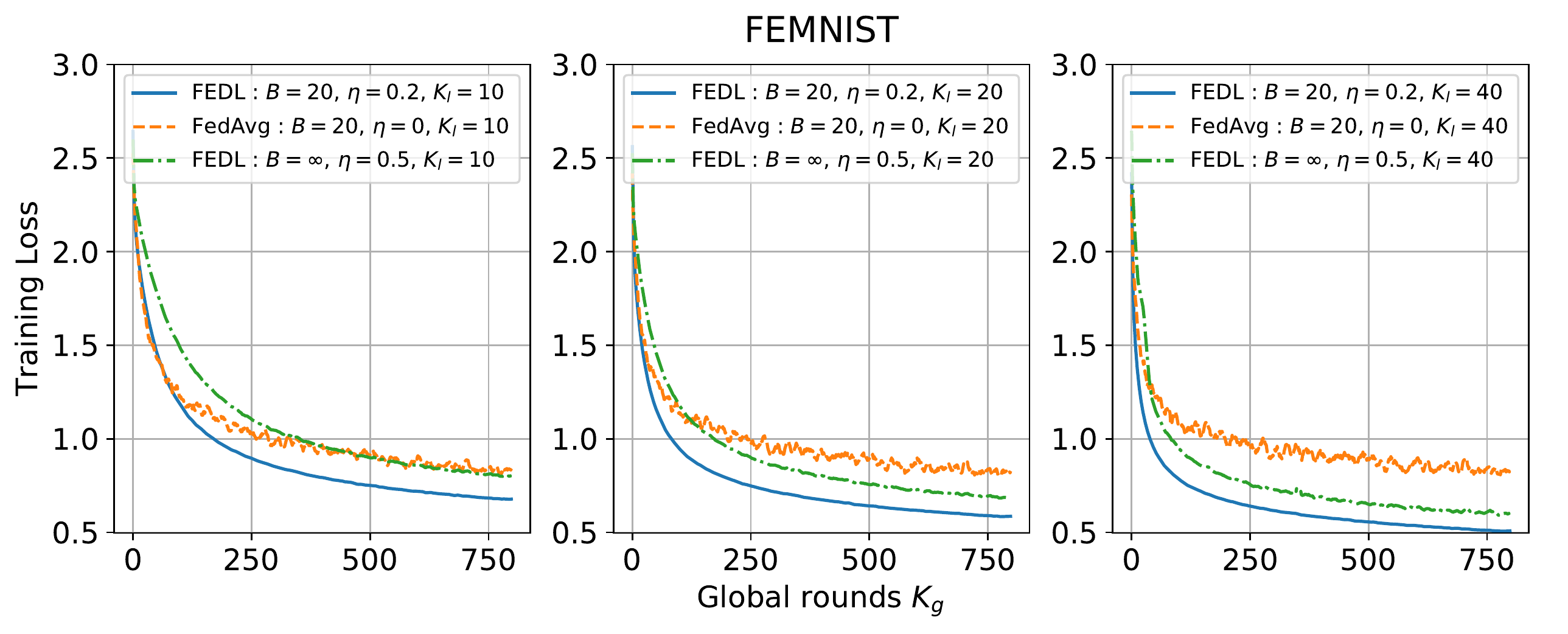}\\
		\caption{Effect of increasing local computation time on the convergence of \FL ($N = 100$, $S = 10$, $K_g = 800$).}
		\label{F:nist_local_com}
	\end{figure}
	
	This section will validate the \FL's learning performance in a heterogeneous network. The experimental results show that \FL gains performance improvement from the vanilla FedAvg \cite{mcmahanCommunicationEfficientLearningDeep2017} in terms of training loss convergence rate and test accuracy in various settings. All codes and data are published on GitHub \cite{dinhCharlieDinhFEDLPytorch2020}.
	
		\begin{figure}[t!]
	\centering
			\begin{subfigure}{0.45\linewidth}
		\includegraphics[width=1.\linewidth]{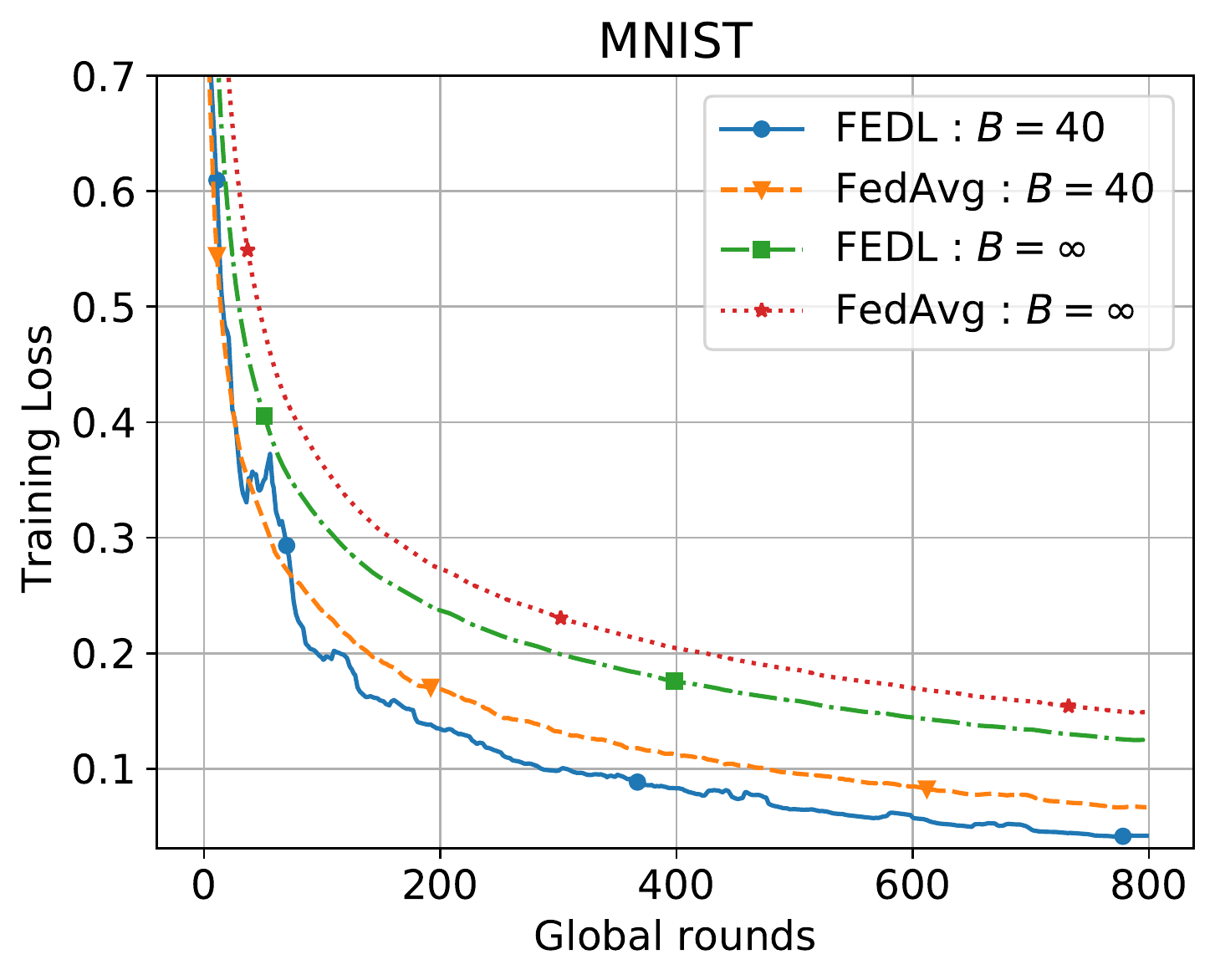}
	\end{subfigure}
	\begin{subfigure}{0.45\linewidth}
		\includegraphics[width=1.\linewidth]{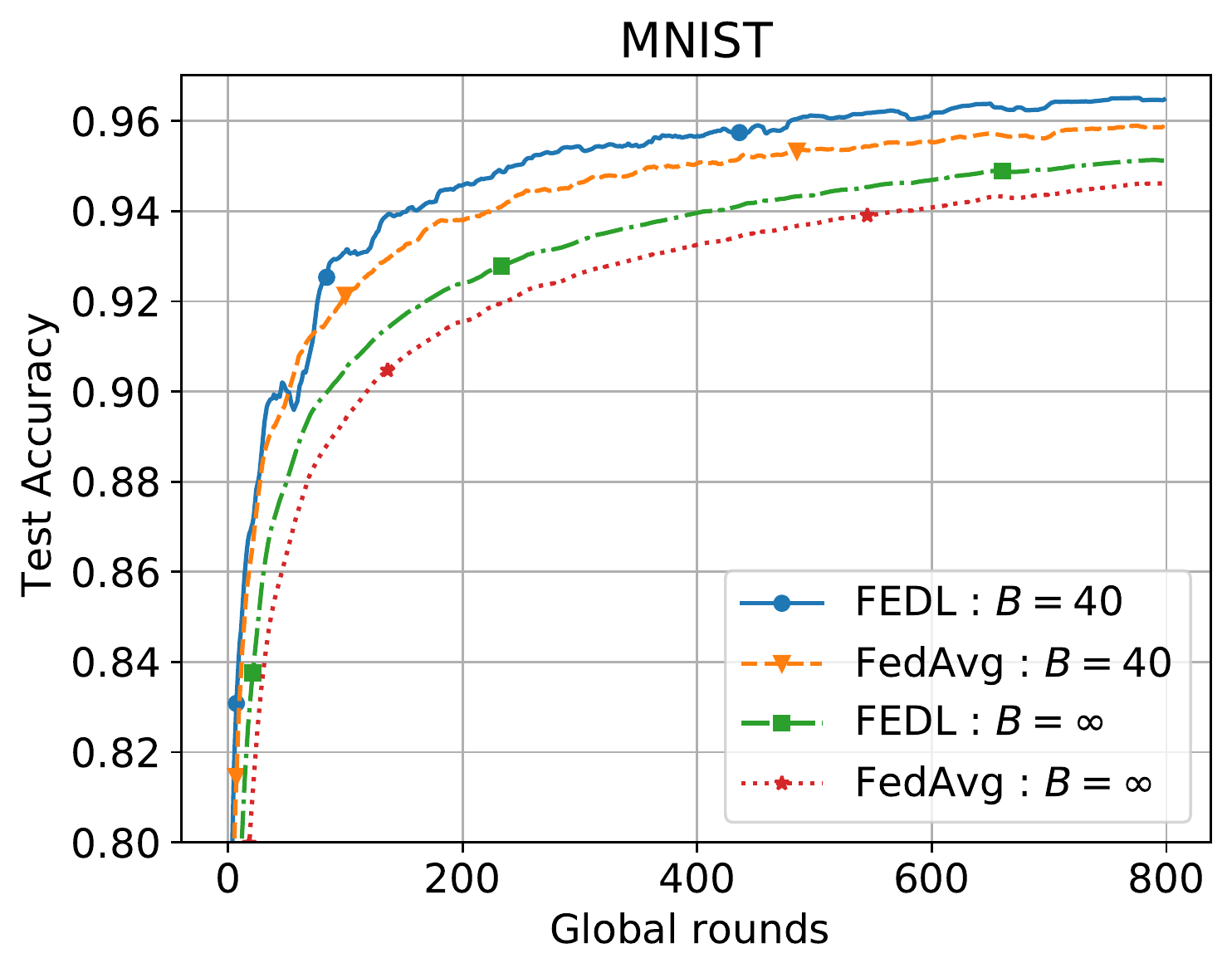}
	\end{subfigure}

		\caption{\FL's performance on non-convex (MNIST dataset).}
		\label{F:mnist_nonconvex}
	\end{figure}
	

	\textbf{Experimental settings:}
	
	In our setting, the performance of \FL is examined by both classification and regression tasks. The regression task uses linear regression model with mean square error loss function on a synthetic dataset while the classification task uses multinomial logistic regression model with cross-entropy error loss function on real federated datasets (MNIST  \cite{lecunGradientbasedLearningApplied1998}, FEMNIST \cite{caldasLEAFBenchmarkFederated2018}). The loss function for each UE is given below:\\

\begin{enumerate}
	\item Mean square error loss for linear regression (synthetic dataset):
		{\small\begin{align*}
			F_n (w)  = \frac{1}{D_n} \sum_{ (x_i,y_i) \in \Cal{D}_n} (\innProd{x_i}{w} - y_i)^2. 
			\end{align*}}
	
	\item Cross-entropy loss for multinomial logistic regression with $C$ class labels (MNIST and FEMNIST datasets):
		{\small
			\begin{align*}
			&F_n (w)   = \frac{-1}{D_n}\biggS{\sum_{i=1}^{D_n}\sum_{c=1}^{C}1_{\{y_i = c\}} \log\frac{\exp(\innProd{x_i}{w_{c}})}{\sum_{j=1}^{C}\exp(\innProd{x_i}{w_{j}})}} \nonumber \\
			& \quad \quad \quad \quad +\frac{\beta}{2}\sum_{c=1}^{C}\norm{w_c}^2. 
			\end{align*}}
	
\end{enumerate}

	{We consider $N = 100$ UEs. To verify that FEDL also works with UE subset sampling, we allow FEDL to randomly sample a number of subset of UEs, denoted by $S$, following a uniform distribution as in FedAvg in each global iteration.} In order to generate datasets capturing the heterogeneous nature of FL, all datasets have different sample sizes based on the power law in \cite{liFederatedOptimizationHeterogeneous2019}. In MNIST, each user contains three of the total of ten labels. FEMNIST is built similar to \cite{liFederatedOptimizationHeterogeneous2019} by partitioning the data in Extended MNIST \cite{cohenEMNISTExtensionMNIST2017}. { For synthetic data, to allow for the non-iid setting, each user's data has the dimension $d = 40$. We  control the value of $\rho$ by using the data generation method similar to that in \cite{liCommunicationEfficientDecentralized2019}, in which user $i$ has data drawn from $\Cal{N}(0,\sigma_i \Sigma)$ where $\sigma_i \thicksim \mathcal{U}(1, 10)$. Here $\Sigma$ is a diagonal covariance matrix with $\Sigma_{ii} = i^{-p}, i \in [1,d]$ and $p = \frac{\log(\rho)}{\log(d)}$, where $\rho$ is considered as multiplicative inverse for the minimum covariance value of $\Sigma$. The number of data samples of each UE is in the ranges $[55,3012]$, $[504,1056]$, and $[500, 5326]$ for MNIST, FEMNIST, and Synthetic, respectively. All datasets are split randomly with 75\% for training and 25\% for testing. Each experiment is run at least 10 times and the average results are reported. We summarized all parameters used for the experiments in Table. \ref{T:parameters}.}
\begin{table}[H]
		\centering
		\begin{tabular}{|c|c|}
			\hline
			Parameters &  Description \\
			\hline
			$\rho$ & Condition number of $F_n(\cdot)'s$ Hessian Matrix,  $\rho = \frac{L}{\beta}$ \\
			$K_g$ & Global rounds for global model update \\
			$K_l$ & Local rounds for local model update  \\
			$\eta$ &Hyper-learning rate\\
			$h_k$ & Local learning rate \\	
			$B$ & Batch Size ($B = \infty$ means full batch size for GD) \\
			\hline
		\end{tabular}
		\caption{Experiment  parameters.} \label{T:parameters}
	\end{table}

\textbf{Effect of the hyper-learning rate on \FL's convergence:} 
{We first verify the theoretical finding by predetermining the value of $\rho$ and observing the impact of changing $\eta$ on the convergence of \FL using a synthetic dataset. In Fig. \ref{F:effects_of_eta}, we examine four different values of $\rho$. As can be seen in the figure, with all value of $\rho$, there were exist small enough values of $\eta$ that allow \FL to converge. We also observe that using the larger value of $\eta$ makes \FL converge faster. In addition, even if \FL allows UEs to solve their local problems approximately, the experiment shows that the gap between the optimal solution and our approach  in Fig. \ref{F:effects_of_eta} is negligible. It is noted that the optimal solution is obtained by solving directly the global loss function (\ref{E:Global_Loss}) as we consider the local loss function at each UE is mean square error loss.}
	
\textbf{Effect of different gradient descent algorithms on \FL's performance: }
As UEs are allowed to use different gradient descent methods to minimize the local problem \eqref{E:Computation}, the convergence of \FL can be evaluated on different optimization algorithms: GD and mini-batch SGD by changing the configuration of the batch size during the local training process. Although our analysis results are based on GD, we also monitor the behavior of \FL using SGD in the experiments for the comparison. While a full batch size is applied for GD, mini-batch SGD is trained with a batch size of 20 and 40 samples. 
{We conducted a grid search on $h_k$ to find the value allowing \FL and  FedAvg to obtain the best performance in terms of accuracy and stability. Fig.~\ref{F:nist_local_batch} demonstrates that \FL outperforms FedAvg on all batch size settings (the improvement in terms of   testing accuracy and training loss are approximately 1.3\% and 9.1\% respectively for the batch size 20, 0.7\% and -0.2\% for the batch size 40, and 0.8\% and 14\% for the full batch size). Besides, \FL is more stable than FedAvg when the small number of devices is sub-sampling randomly during the training process. Even though using larger batch size benefits the stability of both FedAvg and \FL, very large batch size can make the convergence of \FL slow.} However, increasing the value of $\eta$ allows speeding up the convergence of \FL in case of GD. 
	
\textbf{Effect of increasing local computation on convergence time: }
In order to validate the performance of \FL on a different value of local updates $K_l$, in the Fig.~\ref{F:nist_local_com}, we use both mini-batch SGD algorithm with the fixed batch size of 20 and GD for the local update and increase $K_l$ from 10 to 40. 
{ For all $K_l$, even when the learning rate of FedAvg is tuned carefully, \FL in all batch size settings achieve a significant performance gap over FedAvg in terms of training loss and testing accuracy. Also in this experiment, \FL using minibatch outperforms \FL using GD for FEMNIST dataset. While larger $K_l$ does not show the improvement on the convergence of FedAvg, the rise of $K_l$ has an appreciably positive impact on the convergence time \FL. However, the larger $K_l$ requires higher local computation at UEs, which costs the EU's energy consumption.}

\textbf{\FL's performance on non-convex problem:}
Even though the convergence of FEDL is only applicable to strongly convex and smooth loss functions  in theory,  we will show that  FEDL also empirically works well in  non-convex case. We consider a simple non-convex model that has a two-layer deep neural network (DNN) with one hidden layer of size 100 using ReLU activation, an output layer using a softmax function, and the negative log-likelihood loss function.  In Fig.~\ref{F:mnist_nonconvex}, we see that by using a more complexed model, both \FL and FedAvg achieve higher accuracy than the strongly convex case. Although \FL still outperforms FedAvg in case of DNN, the performance improvement is negligible compared to the strongly convex case, and \FL  become less stable when using small mini-batch size.


	
	
	\section{Numerical Results} \label{S:Sim}
	\begin{figure}
		\centering
		\begin{subfigure}{0.49\linewidth} 
			\includegraphics[width=1.\linewidth]{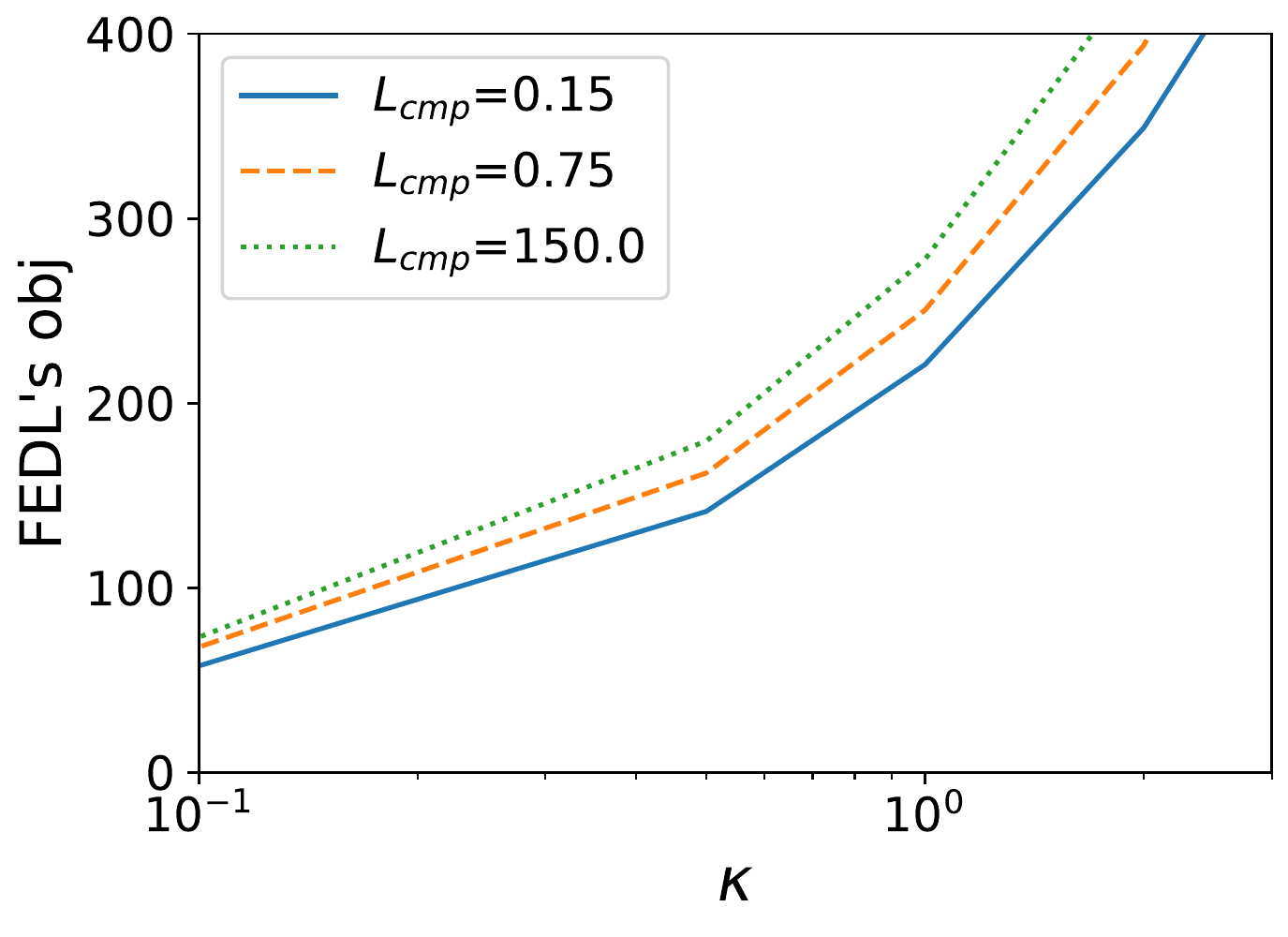}
			\caption{Impact of \Lcomp}
			\label{F:TotalCost_l1}
		\end{subfigure}
		\begin{subfigure}{0.49\linewidth} 
			\includegraphics[width=1.\linewidth]{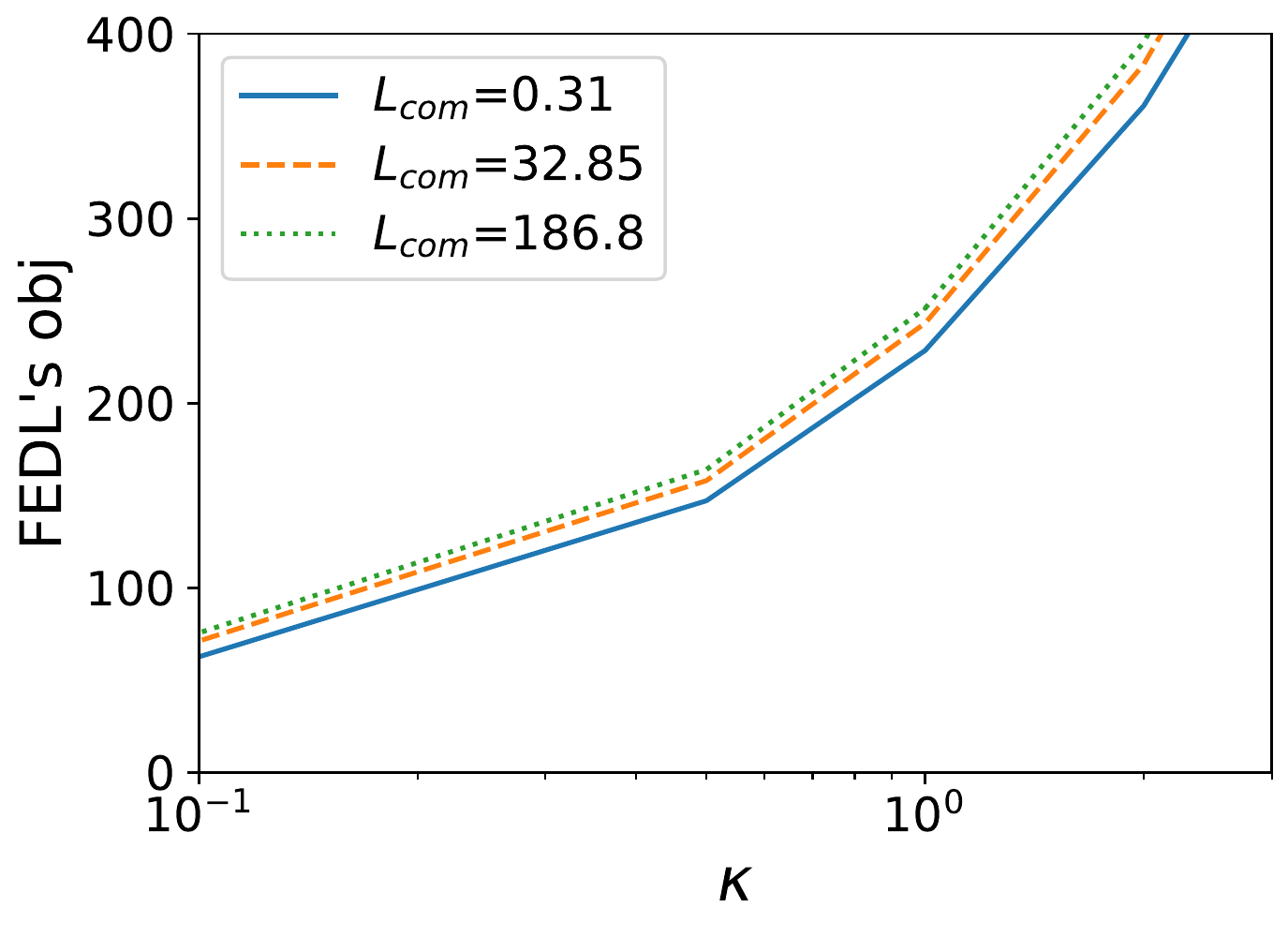}
			\caption{Impact of \Lcomm}
			\label{F:TotalCost_l2}
		\end{subfigure}
		\caption{Impact of UE heterogeneity on  \Opt. } \label{F:TotalCost}
	\end{figure}

	\begin{figure}[!t]
		\centering
		\begin{subfigure}{0.49\linewidth} 
			\includegraphics[width=1.\linewidth]{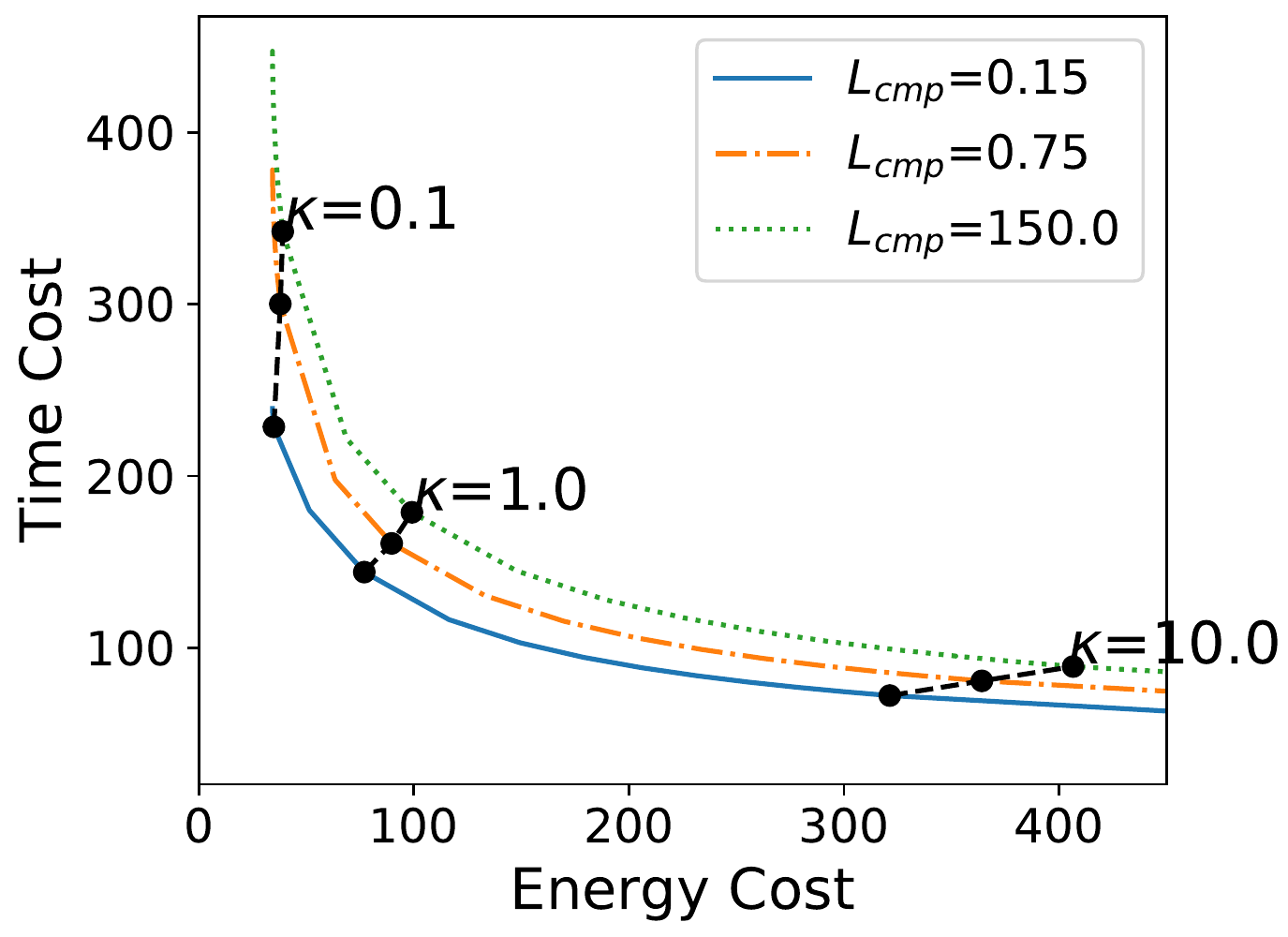}
			\caption{Impact of \Lcomp\ and $\kappa$.}
			\label{F:Pareto_l1}
		\end{subfigure}
		\begin{subfigure}{0.49\linewidth} 
			\includegraphics[width=1.\linewidth]{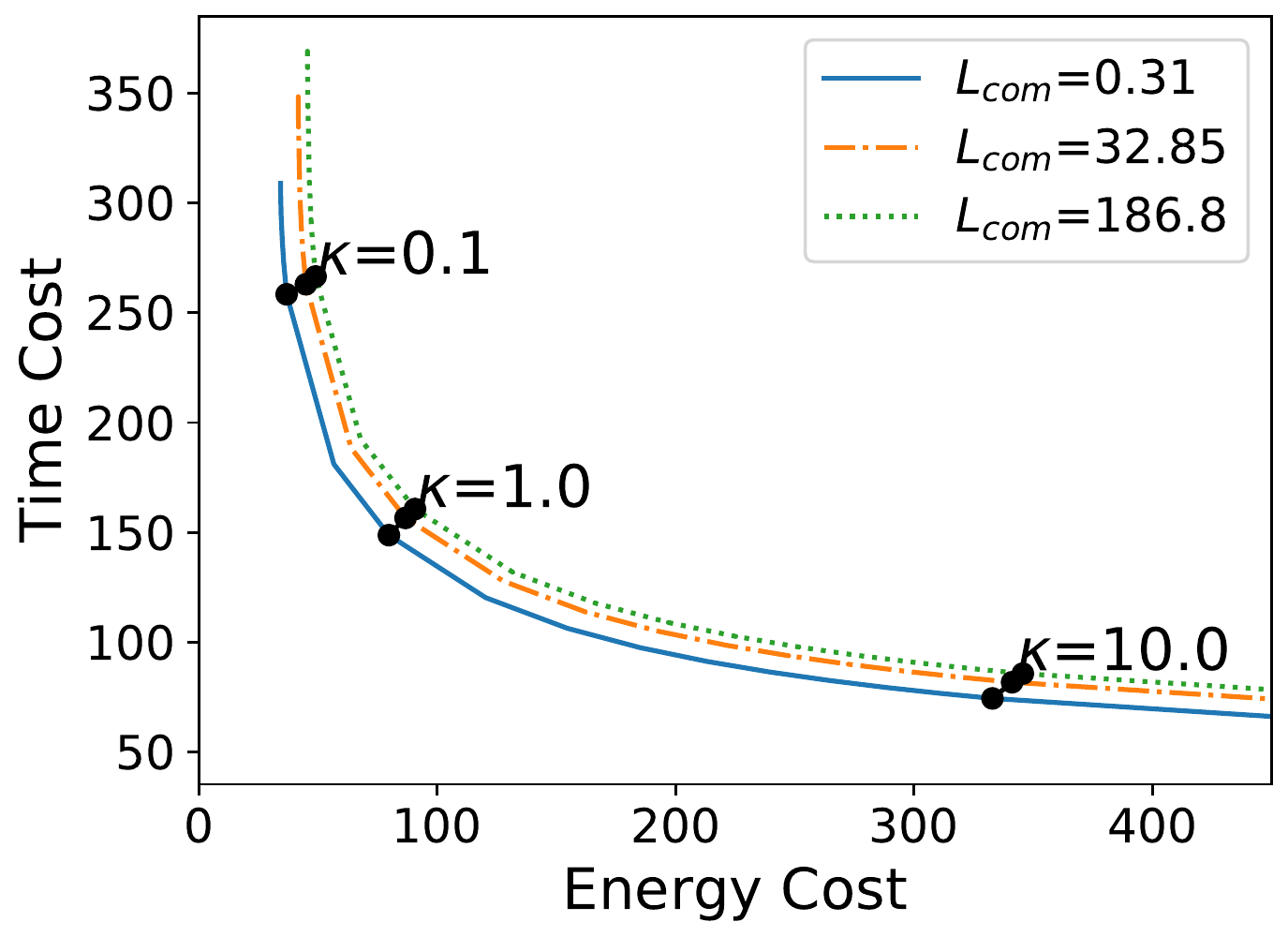}
			\caption{Impact of \Lcomm\ and $\kappa$.}
			\label{F:Pareto_l2}
		\end{subfigure}
		\caption{Pareto-optimal points of  \Opt.} \label{F:Pareto}
	\end{figure}

	In this section, both the communication and computation models follow the same setting as in Fig.~\ref{F:Sub1}, except the number of UEs is increased to 50, and all UEs have the same $f^{max}_n = 2.0$ GHz and $c_n = 20$ cycles/bit. Furthermore, we define two new parameters, addressing the UE heterogeneity regarding to computation and communication phases in \FL, respectively, as $\Lcomp  = \frac{\max_{ n \in \Cal{N}}  \frac{ c_n D_n }{\fMax}} {\min_{ n \in \Cal{N}} \frac{c_n D_n }{ \fMin}}$ and $\Lcomm = \frac{ \max_{ n \in \Cal{N}}  \tauMin}{ \min_{ n \in \Cal{N}}  \tauMax}$.
	We see that higher values of \Lcomp\ and \Lcomm\  indicate higher levels of UE heterogeneity.   The level of heterogeneity is controlled by two different settings. To vary \Lcomp, the training size $D_n$ is generated with the fraction $\frac{D^{min}}{D^{max}} \in  \bigC{1, 0.2, 0.001}$ but the average UE data size is kept at the same value  $7.5$ MB for varying values of  \Lcomp. On the other hand, to vary \Lcomm, the distance between these devices and the edge server is generated such that $\frac{d^{min}}{d^{max}} \in  \bigC{1., 0.2, 0.001}$ but the average distance of all UEs is maintained at  $26$ m. 
	
	
	We first examine the impact of UE heterogeneity. We observe that the high level of UE heterogeneity has negative impact on the \FL system, as illustrated in Figs.~\ref{F:TotalCost_l1} and \ref{F:TotalCost_l2}, such that the total cost is increased with  higher value of \Lcomp\ and \Lcomm\, respectively.
	We next illustrate the Pareto curve in Fig.~\ref{F:Pareto}. This curve shows the trade-off between the conflicting goals of minimizing the  time cost $K(\theta)\Titer\ $ and energy cost  $K(\theta) \Eiter$, in which we can decrease one type of cost yet with the expense of  increasing the other one. This figure also shows that  the Pareto curve of \FL\ is more efficient when the system has low level of UE heterogeneity (i.e., small \Lcomp\ and/or \Lcomm).

	\section{Conclusions} \label{S:Conclusion}
	In this paper, we studied FL, a learning scheme in which the training model is distributed to participating UEs performing training over their local data. Although FL shows vital advantages in data privacy, the heterogeneity across users’ data and UEs' characteristics are still challenging problems. We proposed an effective algorithm without the i.i.d. UEs' data assumption for strongly convex and smooth FL's problems and then characterize the algorithm's convergence. For the wireless resource allocation problem, we embedded the proposed FL algorithm in wireless networks which considers the trace-offs not only between computation and communication latencies but also the FL time and UE energy consumption. Despite the non-convex nature of this problem, we decomposed it into three sub-problems with convex structure before analyzing their closed-form solutions and quantitative insights into problem design. We then verified the theoretical findings of the new algorithm by experiments on Tensoflow with several datasets, and the wireless resource allocation sub-problems by extensive numerical results. In addition to validating the theoretical convergence, our experiments also showed that the proposed algorithm can boost the convergence speed compared to an existing baseline approach.

	\section*{Acknowledgment}
	{{This research is funded by Vietnam National Foundation for Science and Technology Development (NAFOSTED) under grant number 102.02-2019.321. Dr. Nguyen H. Tran and Dr. CS Hong are the corresponding authors. }

	\bibliographystyle{IEEEtran}
	\bibliography{FEDL_Canh,FEDL_journal}
	
	\appendix
	
	\subsection{Review of useful existing results}
	With Assumption~\ref{Assumption} on   $L$-smoothness and $\beta$-strong convexity of $F_n(\cdot)$, according to \cite{nesterovLecturesConvexOptimization2018}[Theorems 2.1.5, 2.1.10, and 2.1.12], we have the following useful inequalities
	{\small
		\vspace{-0.1cm}
		\begin{align}
		2 	L (F_n (w) - F_n (w^{*})) &\geq \norm{\nabla F_n (w)}^2, \forall w \label{E:existing_smooth} \\
		\langle \nabla F_n(w) \!-\! \nabla F_n(w'), w - w' \rangle\! &\geq \!\frac{1}{L} \! \norm{\nabla F_n(w) \!-\! \nabla F_n(w')}^2 \label{E:coercivity}\\ 
		2 	\beta (F_n (w) - F_n (w^{*})) &\leq \norm{\nabla F_n (w)}^2, \forall w \label{E:existing_strong_convex} \\
		\beta \norm{w - w^{*}} &\leq \norm{\nabla F_n (w)}, \forall w, \label{E:existing_strong_convex1} 
		\vspace{-0.1cm}
		\end{align}
	}
	where $w^{*}$ is the solution to problem $\min_{w \in \mathbb{R}^d} F_n(w)$. 
	
	\vspace{-0.4cm}
	
	\subsection{Proof of Lemma~\ref{Lem:local}}
	Due to  $L$-smooth and $\beta$-strongly convex  $J_{n}$, from \eqref{E:existing_smooth} and \eqref{E:existing_strong_convex}, respectively, we have
	{\small
		\vspace{-0.4cm}
		\begin{align*}
		J_{n}^{t} (z_k)  - J_{n}^{t} (z^*) \geq \frac{\norm{\nabla J_{n}^{t} (z_k)}^2}{2 L },  \\
		J_{n}^{t} (z_0)  - J_{n}^{t} (z^*) \leq \frac{\norm{\nabla J_{n}^{t} (z_0)}^2}{2 \beta}. 
		\end{align*}
	}	
	Combining these inequalities with \eqref{E:gradient_based}, and setting $z_0 = w^{t-1}$ and $z_k = w_n^t$, we have 
	{\small
		\vspace{-0.1cm}
		\begin{align*}
		\norm{\nabla J_{n}^{t} (w_n^t)}^2  &\leq c  \frac{L}{\beta } (1 - \gamma)^k  \norm{\nabla J_{n}^{t} (w^{t-1})}^2. 
		\end{align*}
	}
	Since $(1 - \gamma)^k \leq e^{-k \gamma}$, the $\theta$-approximation condition \eqref{E:theta_approximation} is satisfied when
	{\small
		\begin{align*}
		c  \frac{L}{\beta } e^{-k \gamma}  &\leq \theta^2. 
		\end{align*}
	}
	Taking $\log$ both sides of the above, we complete the proof. 
	\vspace{-0.6cm}
	\subsection{Proof of Theorem~\ref{Th:1}}
	We remind the  definition of $J_n^t (w)$ as follows
	\vspace{-0.5cm}
	{\small
		
		\begin{align}
		J_n^t (w)  = F_n(w) + \langle  \eta \nab - \nabla F_n(w^{t-1}),w\rangle. \label{E:J}
		\end{align}
	}
	Denoting $\hat{w}_n^{t}$ the solution to $\min_{w \in \DD{R}^d} J_n^t (w)$, we have
	{\small
		\vspace{-0.1cm}
		\begin{align}
		\nabla J_n^t (w^{t-1}) &= \eta \nab, \label{E:grad_J_past}\\ 
		\!\! \nabla J_n^t (\hat{w}_n^{t})  &=  0 = \nabla F_n(\hat{w}_n^{t})  +  \eta \nab - \nabla F_n(w^{t-1}). \label{E:grad_J_current_n}
		\end{align}
	}
	Since $F(\cdot)$ is also $L$-Lipschitz smooth (i.e., {\small$\norm{ \nabla F (w) - \nabla F (w') } \leq \SumNoLim{n=1}{N} \frac{D_n}{D}\norm{ \nabla F_n (w) - \nabla F_n (w') } \leq L \norm{w - w'}, \forall w, w',$} by using Jensen's inequality and $L$-smoothness, respectively), we have
	{\small \begin{align}
		&F(w_n^t) - F(w^{t-1})  \nonumber \\
		& \leq  \langle \nabla F(w^{t-1}), w_n^t - w^{t-1}\rangle + \frac{L}{2} \norm{w_n^t-w^{t-1}}^2 \nonumber \\
		& =  \langle \nabla F(w^{t-1}) - \nab, w_n^t - w^{t-1}\rangle + \frac{L}{2} \norm{w_n^t-w^{t-1}}^2  \nonumber\\
		&\, \quad + \langle \nab, w_n^t - w^{t-1}\rangle \label{E:add_subtract1} \\
		& \leq  \norm{ \nabla F(w^{t-1}) - \nab} \norm{w_n^t - w^{t-1}}  + \frac{L}{2} \norm{w_n^t-w^{t-1}}^2  \nonumber\\
		&\, \quad + \langle \nab, w_n^t - w^{t-1}\rangle \label{E:Cauchy} \\
		& \!\! \stackrel{(\ref{E:grad_J_current_n})}{=}  \norm{ \nabla F(w^{t-1}) - \nab} \norm{w_n^t - w^{t-1}}  + \frac{L}{2} \norm{w_n^t-w^{t-1}}^2  \nonumber\\
		&\, \quad  -\frac{1}{\eta} \langle \nabla F_n(\hat{w}_n^{t}) - \nabla F_n(w^{t-1}), w_n^t - w^{t-1}\rangle \nonumber \\
		&= \norm{ \nabla F(w^{t-1}) - \nab} \norm{w_n^t - w^{t-1}}  + \frac{L}{2} \norm{w_n^t-w^{t-1}}^2  \nonumber\\
		&\, \quad  -\frac{1}{\eta} \langle \nabla F_n(\hat{w}_n^{t}) - \nabla F_n(w_n^t), w_n^t - w^{t-1}\rangle \nonumber \\		
		&\, \quad  -\frac{1}{\eta} \langle \nabla F_n(w_n^t) - \nabla F_n(w^{t-1}), w_n^t - w^{t-1}\rangle \label{E:add_subtract2} \\		
		&\leq  \norm{ \nabla F(w^{t-1}) - \nab} \norm{w_n^t - w^{t-1}}  + \frac{L}{2} \norm{w_n^t-w^{t-1}}^2  \nonumber\\
		&\, \quad  + \frac{L}{\eta} \norm{\hat{w}_n^t - w_n^{t}} \norm{w_n^t - w^{t-1}} \label{E:Cauchy2} \\		
		&\, \quad  -\frac{1}{\eta} \langle \nabla F_n(w_n^t) - \nabla F_n(w^{t-1}), w_n^t - w^{t-1}\rangle \nonumber \\	
		&\!\! \stackrel{(\ref{E:coercivity})}{\leq}  \norm{ \nabla F(w^{t-1}) - \nab} \norm{w_n^t - w^{t-1}}  + \frac{L}{2} \norm{w_n^t-w^{t-1}}^2  \nonumber\\
		&\, \quad  + \frac{L}{\eta} \norm{\hat{w}_n^t - w_n^{t}} \norm{w_n^t - w^{t-1}} \nonumber \\		
		&\, \quad  - \frac{1}{\eta L  } \norm{\nabla F_n(w_n^{t}) - \nabla F_n(w^{t-1})}^2,  \label{E:convergence1}
		\end{align} 
	}
	where \eqref{E:add_subtract1} is by adding and subtracting $\nab$, \eqref{E:Cauchy} is by Cauchy-Schwarz inequality, \eqref{E:add_subtract2} is by adding and subtracting $\nabla F_n(w_n^{t})$, \eqref{E:Cauchy2} is by using Cauchy-Schwarz inequality and $L$-Lipschitz smoothness of $F_n(\cdot)$. The next step is to bound the norm terms in the R.H.S of \eqref{E:convergence1} as follows:  
	\begin{itemize}
		\item  First, we have
		{\small
			\vspace{-0.2cm}
			\begin{align}
			& \norm{\hat{w}_n^t - w^{t-1}} \stackrel{(\ref{E:existing_strong_convex1})}{\leq}  \frac{1}{\beta}\norm{\nabla J_n^t (w^{t-1})} \stackrel{(\ref{E:grad_J_past})}{=} \frac{\eta}{\beta}\norm{\nab}. \label{E:dist1}
			\end{align}
		}
		\vspace{-0.4cm}
		\item Next, 
		{\small
			\vspace{-0.2cm}
			\begin{align}	
			\norm{\hat{w}_n^t - w_n^{t}} &  \stackrel{(\ref{E:existing_strong_convex1})}{\leq}  \frac{1}{\beta}\norm{\nabla J_n^t (w_n^t)}  \stackrel{(\ref{E:theta_approximation})}{\leq} \frac{\theta}{\beta}\norm{\nabla J_n^t (w^{t-1})}  \nonumber \\
			&\stackrel{(\ref{E:grad_J_past})}{=}   \frac{ \theta \eta}{\beta}\norm{\nab}. \label{E:dist2}
			\end{align}
		}
		\vspace{-0.4cm}
		\item 	Using triangle inequality, \eqref{E:dist1}, and \eqref{E:dist2}, we have
		{\small
			\vspace{-0.2cm}
			\begin{align}
			\norm{w_n^t - w^{t-1}}  &\leq   \norm{ w_n^{t} - \hat{w}_n^t}  + \norm{\hat{w}_n^t - w^{t-1}} \nonumber\\
			& \leq (1 + \theta)\frac{\eta}{\beta} \norm{\nab}. \label{E:dist3}
			\end{align}
		}
		\vspace{-0.4cm}
		\item We also have
		{\small
			\vspace{-0.2cm}
			\begin{align}
			\!\!\norm{\nabla F_n(w_n^{t}) - \nabla F_n(w^{t-1})} \!&\stackrel{(\ref{E:J})}{=} \norm{\nabla J_n^t (w_n^t) - \nabla J_n^t (w^{t-1})} \nonumber\\
			&  \geq \norm{\nabla J_n^t (w^{t-1})} - \norm{\nabla J_n^t (w_n^t)} \nonumber \\
			&  \stackrel{(\ref{E:theta_approximation})}{\geq}  (1-\theta)\norm{\nabla J_n^t (w^{t-1})} \nonumber \\
			&  \stackrel{(\ref{E:grad_J_past})}{=}   (1-\theta) \eta \norm{\nab}.  \label{E:dist4}
			\end{align}
		}
		\vspace{-0.4cm}
		\item  By definitions of $\nabla F(.)$ and $\nab$, we have
		{\small
			\vspace{-0.2cm}
			\begin{align}
			\!\!\!\!	\norm{ \nabla F(w^{t-1}) - \nab} &=  \norm{ \SumLim{n=1}{N} p_n  \bigP{\nabla F_n(w_n^{t}) - \nabla F_n(w^{t-1})} } \nonumber \\
			& \leq  \SumLim{n=1}{N} p_n \norm{\nabla F_n(w_n^{t}) - \nabla F_n(w^{t-1}) } 
			\label{E:dist5} 
			\end{align}	
			\vspace{-0.8cm}
			\begin{align}
			\qquad \qquad & \leq  \SumLim{n=1}{N} p_n L \norm{w_n^{t} - w^{t-1} } 	\label{E:dist6} \\
			\qquad \qquad & \leq (1 + \theta) \eta \frac{ L}{\beta} \norm{\nab}, \label{E:dist7}
			\end{align}
		}
		where \eqref{E:dist5}, \eqref{E:dist6} and \eqref{E:dist7} are obtained using Jensen's inequality, $L$-Lipschitz smoothness, and \eqref{E:dist3}, respectively. 
		\item Finally,  we have
	{	{\small
			\begin{align}
			\norm{\nabla F(w^{t-1}) }^2 &\leq 2 \norm{ \nab - \nabla F(w^{t-1}) }^2 + 2 \norm{\nab }^2 \label{E:dist8} \\
			\qquad &\stackrel{(\ref{E:dist7})}{\leq} 2 (1 + \theta)^2 \eta^2 \rho^2 \norm{\nab}^2 + 2 \norm{\nab }^2 \nonumber
			\end{align}
		}
	
		which implies
		{\small
			\begin{align}
			\norm{\nab}^2 &\geq \frac{1}{ 2 (1 + \theta)^2 \eta^2 \rho^2 +2 } \norm{\nabla F(w^{t-1}) }^2, \label{E:dist9}
			\end{align}
		}
	}
	where \eqref{E:dist8} comes from the fact that $\norm{x + y}^2 \leq 2 \norm{x}^2 + 2 \norm{y}^2$ for any two vectors $x$ and $y$.
	\end{itemize}
	\vspace{0.8em}
		Defining 
	{\small 
		\begin{align}
		Z &\defeq \frac{\eta(-2(\theta-1)^2+ (\theta+1)\theta(3\eta+2)\rho^2+(\theta+1)\eta\rho^2)}{2\rho} < 0 \nonumber 
		\end{align}
	}
	and	substituting \eqref{E:dist1}, \eqref{E:dist3}, \eqref{E:dist4}, and \eqref{E:dist7} into \eqref{E:convergence1}, we have
	{
	{\small 
		\begin{align}
		&\!\!\!\!\!\!\!\!\!\!\!\!\!\!\!\!\!\!\!\!\!\!\!\!\!\!\!\!\!\!\!\!\!\!\!\!\! \!\!\!\!\!\!\!\!\!\!\!  F(w_n^t) - F(w^{t-1})  \nonumber \\
		&\!\!\!\!\!\!\!\!\!\!\!\!\!\!\!\!\!\!\!\!\!\!\!\!\!\!\!\!\!\!\!\!\!\!\!\!\!\!\!\!\!\!\!\!\!\!\!\!  \leq \frac{Z}{\beta} \norm{\nab}^2 \nonumber \\
		&\!\!\!\!\!\!\!\!\!\!\!\!\!\!\!\!\!\!\!\!\!\!\!\!\!\!\!\!\!\!\!\!\!\!\!\!\!\!\!\!\! \!\!\!\!\!\!\!  \stackrel{(\ref{E:dist9})}{\leq} \frac{Z}{2\beta\bigP{(1+\theta)^2\eta^2\rho^2 + 1}}\norm{\nabla F(w^{t-1})}^2 \nonumber
		\end{align}	
		\vspace{-0.4cm}
		\begin{align}
		&\! \stackrel{(\ref{E:existing_strong_convex})}{\leq} -  \frac{-Z}{\bigP{(1+\theta)^2\eta^2\rho^2 + 1}}   (F(w^{t-1}) - F(w^{*}))  \label{E:Convergence11}\\
		&\! = - \frac{\eta(2(\theta-1)^2- (\theta+1)\theta(3\eta+2)\rho^2-(\theta+1)\eta\rho^2)}{2\rho\bigP{(1+\theta)^2\eta^2\rho^2 + 1}}(F(w^{t-1}) - F(w^{*})) \nonumber\\
		&\!\! \stackrel{(\ref{E:Theta})}{=}  - \Theta \, (F (w^{t-1}) - F(w^{*})).   \label{E:Convergence2}
		\end{align}
	}

}
	By subtracting $F(w^{*})$ from both sides of \eqref{E:Convergence2},  we have
	{\small 
		\begin{align}
		F(w_n^t) - F(w^{*}) \leq (1 - \Theta) \, \bigP{F (w^{t-1}) - F(w^{*})}, \forall n. \label{E:Convergence3}
		\end{align}
	}
	Finally, we obtain
	{\small 
		\begin{align}
		F(w^t) - F(w^{*}) &\leq \SumNoLim{n=1}{N} p_n \bigP{F(w_n^t) - F(w^{*})} \label{E:Convergence31}\\
		&\stackrel{(\ref{E:Convergence3})}{\leq } (1 - \Theta) \, (F (w^{t-1}) - F(w^{*})), \label{E:Convergence4}
		\end{align}
	}
	where \eqref{E:Convergence31} is due to the convexity of $F(\cdot)$.

	\subsection{Proof of Lemma~\ref{L:1}}
	The convexity of \SubOne can be shown by its strictly convex objective in \eqref{E:Ecomp} and its constraints determine a convex set. Thus, the global optimal solution of \SubOne can be found using KKT condition \cite{boydConvexOptimization2004}.  In the following, we first provide the KKT condition of \SubOne, then show that solution in Lemma~\ref{L:1} satisfy this condition. 
	
	The Lagrangian of \SubOne is 
	{\small
		\begin{align*}
		L_1 &= \SumN \BigS{\Ecomp  +  \lambda_n ( \frac{c_n D_n}{f_n} - \Tcomp )   \nonumber\\  
			&+  \mu_n( f_n - \fMax) - \nu_n( f_n - \fMin)} + \kappa  \Tcomp
		\end{align*}
	}
	where $\lambda_n, \mu_n, \nu_n$ are non-negative dual variables with their optimal values denoted by  $\lambda_n^*, \mu_n^*, \nu_n^*$, respectively. Then the KKT condition is as follows:
	{\small
		\begin{align}
		\pder[L]{f_n} = \pder[\Ecomp ]{\tau_n} - \lambda_n \frac{c_n D_n}{f_n^2}  + \mu_n - \nu_n &= 0, \; \forall n \label{E:KKT1'} 	\\
		\pder[L]{\Tcomp} = \kappa - \SumN \lambda_n &= 0,  \label{E:KKT2'}\\
		\mu_n( f_n - \fMax) &= 0, \; \forall n \label{E:KKT3'}\\
		\nu_n( f_n - \fMin) &= 0, \; \forall n \label{E:KKT4'}\\
		\lambda_n (\frac{c_n D_n}{f_n} - \Tcomp ) &= 0. \; \forall n \label{E:KKT5'}
		\end{align}
	}
	
	Next, we will show that the optimal solution according to KKT condition is also the same as that provided by Lemma~\ref{L:1}. To do that, we observe that the existence of \None, \Ntwo, \Nthree\ and their respective \TnOne, \TnTwo, \TnThree\ produced by Algorithm~\ref{Alg1}  depends on $\kappa$. Therefore, we will construct the ranges of $\kappa$ such that there exist three subsets  $\None', \Ntwo', \Nthree'$ of UEs satisfying KKT condition and having the same solution as that in  Lemma~\ref{L:1} in the following cases. 
	
	\begin{enumerate}[a)]	
		\vspace{0.5cm}
		\item $\TcompOpt=\TnOne \geq \max\bigC{\TnTwo, \TnThree}$: This happens when $\kappa$ is large enough so that the condition in line \ref{Alg1:ifN1} of Algorithm~\ref{Alg1}  satisfies because \TnThree is decreasing when $\kappa$ increase.  Thus we consider  $\kappa  \geq \SumN \alpha_n {(\fMax)}^3$  (which ensures \None\  of Algorithm~\ref{Alg1} is non-empty). 
		
		From \eqref{E:KKT2'}, we have 
		{\small
			\begin{align}
			\kappa  = \SumN \lambda_n^*, \label{E:SumLamb}
			\end{align}
		}
		thus  $\kappa$ in this range can guarantee  a non-empty set  $\None' = \{n | \lambda_n^* \geq  \alpha_n {(\fMax)}^3\}$ such that
		{\small
			\begin{align*}
			\pder[\Ecomp (f_n^*) ]{f_n} - \lambda_n^* \frac{c_n D_n}{{f_n^*}^2}  \leq 0, \forall n \in \None': f_n^* \leq  \fMax. 
			\end{align*}
		}
		Then from \eqref{E:KKT1'} we must have $\mu_n^* - \nu_n^* \geq 0$, thus, according to \eqref{E:KKT3'}  $f_n^* = \fMax,  \, \forall n \in \None'$. From \eqref{E:KKT5'}, we see that $\None'= \{n: \frac{c_n D_n}{\fMax} = \TcompOpt\}$. Hence, by the definition in \eqref{E:constr_comp}, 
		{\small
			\begin{align}
			\TcompOpt = \max_{n \in \Cal{N}} \frac{c_n D_n}{\fMax}.  \label{E:TcompL1}
			\end{align}
		}
		
		On the other hand, if there exist a non-empty set $ \Ntwo' = \{n | \lambda_n^* = 0 \}$, it must be due to
		{\small
			\begin{align*}
			\frac{c_n D_n}{\fMin} \leq \TcompOpt, \; \forall n \in \Ntwo'
			\end{align*}
		}
		according to \eqref{E:KKT5'}. In this case, from  \eqref{E:KKT1'} we must have $\mu_n^* - \nu_n^* \leq 0 \Rightarrow f_n^* = \fMin, \; \forall n \in \Ntwo'$. 
		
		Finally, if there exists UEs with $\frac{c_n D_n}{\fMin} > \TcompOpt$ and $\frac{c_n D_n}{\fMax} < \TcompOpt$, they will belong to the set  $\Nthree' = \{n | \alpha_n {(\fMin)}^3 < \lambda_n^* < \alpha_n {(\fMax)}^3\}$ such that $\fMin < f_n^* < \fMax$. According to \eqref{E:KKT5'}, $\TcompOpt$ must be the same for all $n$ with $\lambda_n^* > 0$ , we obtain
		{\small
			\begin{align}
			f_n^* &= \frac{c_n D_n}{\TcompOpt} = \frac{c_n D_n}{\max_n \frac{c_n D_n}{\fMax}},\; \forall n \in \Nthree'. \label{E:case_a}
			\end{align}
		}
		
		In summary, we have
		{\small
			\begin{align*}	
			f_n^* &= \begin{cases} 
			\fMax,   &\forall {n \in \None'} \\
			\fMin,   &\forall {n \in \Ntwo'} \\
			\frac{c_n D_n}{\TcompOpt},   &\forall {n \in \Nthree'} 
			\end{cases}
			\end{align*}
		}
		with \TcompOpt\ determined in \eqref{E:TcompL1}.

		\item  $\TcompOpt=\TnTwo > \max\bigC{\TnOne, \TnThree}$: This happens when $\kappa$ is small enough such that the condition in line \ref{Alg1:while} of Algorithm~\ref{Alg1} satisfies.  In this case, $\None$ is empty and  \Ntwo\ is non-empty according to line~\ref{Alg1:N2add} of this algorithm. Thus we consider  $ \kappa \leq \SumN \alpha_n {(\fMin)}^3$. 
		
		Due to the considered small $\kappa$ and \eqref{E:SumLamb},  there must exist a non-empty set $\Ntwo' =  \bigC{n \,:\, \lambda_n^* \leq  \alpha_n {(\fMin)}^3}$ such that
		{\small
			\begin{align*}
			\pder[\Ecomp (f_n^*) ]{f_n} - \lambda_n^* \frac{c_n D_n}{{f_n^*}^2} \geq  0, \forall n \in \Ntwo': f_n^* \geq \fMin. 
			\end{align*}
		}
		Then, from \eqref{E:KKT1'} we must have $\mu_n^* - \nu_n^* \leq  0$, and thus from \eqref{E:KKT3'},  $f_n^* = \fMin, \forall n \in \Ntwo'$.  Therefore,   by the definition \eqref{E:constr_comp} we have 
		{\small
			\begin{align*}
			\TcompOpt = \max_{n \in \Ntwo'} \frac{c_n D_n}{\fMin}.
			\end{align*}
		}
		
		If we further restrict $\kappa \leq \min_{n \in \Cal{N}} \alpha_n {(\fMin)}^3$, then we see that $\Ntwo' = \Cal{N}$, i.e., $f_n^* = \fMin, \; \forall n \in \Cal{N}$. On the other hand, if we consider $\SumN \alpha_n {(\fMin)}^3 < \kappa \leq \min_{n \in \Cal{N}} \alpha_n {(\fMin)}^3$, then there may exist UEs with $\frac{c_n D_n}{\fMin} > \TcompOpt$ and $\frac{c_n D_n}{\fMax} < \TcompOpt$, which will belong to the set $\Nthree' = \{n \,|\, \alpha_n {(\fMin)}^3 < \lambda_n^* < \alpha_n {(\fMax)}^3\}$ such that $\fMin < f_n^* < \fMax, \forall n \in \Nthree'$. 	Then, according  to \eqref{E:KKT5'}, $\TcompOpt$ must be the same for all $n$ with $\lambda_n^* > 0$ , we obtain
		{\small
			\begin{align}
			f_n^* &= \frac{c_n D_n}{\TcompOpt} = \frac{c_n D_n}{\max_n \frac{c_n D_n}{\fMin}},\; \forall n \in \Nthree'. \label{E:case_b}
			\end{align}
		}
		In summary, we have
		{\small
			\begin{align*}	
			f_n^* &= \begin{cases} 
			\frac{c_n D_n}{\TcompOpt},   &\forall {n \in \Nthree'}  \\
			\fMin,   &\forall {n \in \Ntwo'}
			\end{cases}
			\end{align*}
		}
		
		\item $\TcompOpt=\TnThree > \max\bigC{\TnOne, \TnTwo}$: This happens when $\kappa$ is between a range such that  in Algorithm~\ref{Alg1}, the condition at line \ref{Alg1:while} is violated at some round, while the condition at line~\ref{Alg1:ifN1} not satisfied.  In this case, $\None$ is empty and  \Nthree\ is non-empty according to line~\ref{Alg1:N2add} of this algorithm. Thus we consider  $\SumN \alpha_n {(\fMin)}^3 < \kappa  <  \SumN \alpha_n {(\fMax)}^3$. 
		
		With this range of $\kappa$, and \eqref{E:SumLamb}, there must exist a non-empty set $\Nthree' = \{n \,|\, \alpha_n {(\fMin)}^3 < \lambda_n^* < \alpha_n {(\fMax)}^3\}$ such that $\fMin < f_n^* < \fMax, \forall n \in \Nthree'$. Then, from  \eqref{E:KKT1'} we have $\mu_n^* - \nu_n^* = 0$ and the following equation
		{\small
			\begin{align*}
			\pder[\Ecomp (f_n) ]{f_n} - \lambda_n^* \frac{c_n D_n}{{f_n}^2}  = 0 
			\end{align*}
		}
		has its solution $f_n^*= \BigP{\frac{ \lambda_n^*}{\alpha_n}}^{1/3},  \forall n \in \Nthree'$. 
		Furthermore, from \eqref{E:KKT5'}, we have 
		{\small
			\begin{align}	
			\frac{c_n D_n}{f_n^*} &= c_n D_n \BigP{\frac{\alpha_n}{\lambda_n^*}}^{1/3} = \TcompOpt, \quad \forall n \in \Nthree'. \label{E:TN3}
			\end{align}
		}

		Combining \eqref{E:TN3} with \eqref{E:SumLamb}, we have
		{\small
			\begin{align*}
			\TcompOpt &= \biggP{\frac{\SumNoLim{n \in \Nthree'}{} \alpha_n (c_n D_n)^3 }{\kappa }}^{1/3}. 
			\end{align*}
		}
		
		On the other hand, if there exist a non-empty set $ \Ntwo' = \{n | \lambda_n^* = 0 \}$, it must be due to
		{\small
			\begin{align*}
			\frac{c_n D_n}{\fMin} <  \TcompOpt, \;  \forall n \in \Ntwo'
			\end{align*}
		}
		according to \eqref{E:KKT5'}. From \eqref{E:KKT1'} we must have $\mu_n^* - \nu_n^* \leq 0 \Rightarrow f_n^* = \fMin, \; \forall n \in \Ntwo'$. In summary, we have
		{\small
			\begin{align*}	
			f_n^* &= \begin{cases} 
			\frac{c_n D_n}{\TcompOpt},   &\forall {n \in \Nthree'} \\
			\fMin,   &\forall {n \in \Ntwo'}
			\end{cases}
			\end{align*}
		}
		
	\end{enumerate}
	
	Considering all cases above, we see that the solutions characterized by KKT condition above are exactly the same as those provided in Lemma~\ref{L:1}. 
	
	\subsection{Proof of Lemma \ref{L:2}}
	According to \eqref{E:Ecomm} and \eqref{E:p_func_of_tau}, the objective of \SubTwo is the sum of perspective functions of convex  and linear functions, and its constraints determine a convex set; thus \SubTwo is a convex problem that can be analyzed using KKT condition \cite{boydConvexOptimization2004}. 
	
	The Lagrangian of \SubTwo is 
	{\small
		\begin{align*}
		&L_2 = \SumN \Ecomm (\tau_n) +  \lambda (\SumN \tau_n - \Tcomm ) \\ \nonumber
		&+ \SumN \mu_n( \tau_n - \tauMax) - \SumN \nu_n( \tau_n - \tauMin)  + \kappa \Tcomm
		\end{align*}
	}
	where $\lambda, \mu_n, \nu_n$ are non-negative dual variables. Then the KKT condition is as follows:
	{\small
		\begin{align}
		\pder[L]{\tau_n} = \pder[\Ecomm ]{\tau_n} + \lambda + \mu_n - \nu_n &= 0, \; \forall n \label{E:KKT1} 	\\
		\pder[L]{\Tcomm} = \kappa - \lambda &= 0,	\label{E:KKT2} \\
		\mu_n( \tau_n - \tauMax) &= 0, \; \forall n \label{E:KKT3}\\
		\nu_n( \tau_n - \tauMin) &= 0, \; \forall n \label{E:KKT4}\\
		\lambda (\SumN \tau_n - \Tcomm ) &= 0. \label{E:KKT5}
		\end{align}
	}
	From \eqref{E:KKT2}, we see that $\lambda^* = \kappa$. Let $x \defeq \frac{s_n}{\tau_n B}$, we first consider the equation
	\newline
	{\small
		\begin{align*}
		&\pder[\Ecomm ]{\tau_n} + \lambda^* = 0	 \Leftrightarrow \frac{N_0}{\bar{h}_n} \bigP{ e^x - 1 - x e^x} = - \lambda^* = -\kappa 	\\
		&\Leftrightarrow e^x (x - 1) = \kappa N_0^{-1} \bar{h}_n - 1 	\Leftrightarrow e^{x-1} (x - 1) = \frac{\kappa N_0^{-1} \bar{h}_n - 1}{e}  	\\
		&\Leftrightarrow x = 1 + {W \BigP{ \frac{ \kappa N_0^{-1} \bar{h}_n - 1}{e} }}	\Leftrightarrow \tau_n = g_n(\kappa) = \frac{s_n/B} {1 +{W \bigP{ \frac{ \kappa N_0^{-1} \bar{h}_n - 1}{e} }}}. 
		\end{align*}
	}
	

	
	Because $W(\cdot)$ is strictly increasing when $W(\cdot)>-\ln2$,  $g_n (\kappa)$ is strictly decreasing and positive, and so is its inverse function 
	{\small
		\begin{align*}
		g_n^{-1}(\tau_n) = - \frac{\partial \Ecomm (\tau_n)}{\partial \tau_n}. 
		\end{align*}
	}
	
	Then we have following cases
	\begin{enumerate}[a)]
		\item If  $g_n(\kappa) \leq  \tauMin \Leftrightarrow \kappa \geq  g_n^{-1}(\tauMin):$
		{\small
			\begin{align*}
			\kappa =\lambda^*  \geq g_n^{-1}(\tauMin) \geq - \frac{\partial \Ecomm}{\partial \tau_n}\Bigr|_{\substack{\tauMin \leq \tau_n}}. 
			\end{align*}
		}
		Thus, according to \eqref{E:KKT1},  $\mu_n^* - \nu_n^* \leq 0$. Because  both $\mu_n^*$ and $\nu_n^*$ cannot be positive, we  have  $\mu_n^* = 0$ and $\nu_n^* \geq  0$.  Then we consider two cases of $\nu^*$: a)  $\nu_n^* > 0$, from \eqref{E:KKT4}, $\tau_n^* = \tauMin$, and b) $\nu_n^* = 0$, from \eqref{E:KKT1}, we must have $\kappa =  g_n^{-1}(\tauMin)$, and thus $\tau_n^* = \tauMin$. 
		
		\item If $g_n(\kappa) \geq  \tauMax \Leftrightarrow \kappa \leq  g_n^{-1}(\tauMax)$, then we have
		{\small
			\begin{align*}
			\kappa =\lambda^* \leq g_n^{-1}(\tauMax) \leq - \frac{\partial \Ecomm}{\partial \tau_n}\Bigr|_{\substack{ \tau_n \leq \tauMax}}. 
			\end{align*}
		}
		Thus, according to \eqref{E:KKT1},   $\mu_n^* - \nu_n^* \geq 0$, inducing  $\nu_n^* = 0$ and $\mu_n^* \geq 0$.  
		With similar reasoning as above, we have  $\tau_n^* = \tauMax$.	
		
		\item If  $ \tauMin < g_n (\kappa) < \tauMax \Leftrightarrow g_n^{-1}(\tauMax) < \kappa < g_n^{-1}(\tauMin)$, then from  \eqref{E:KKT3} and \eqref{E:KKT4}, we must have $\mu_n^* = \nu_n^* = 0$, with which and \eqref{E:KKT1} we have
		{\small
			\begin{align*}
			\tau_n^* = g_n (\kappa). 		
			\end{align*}
		}
	\end{enumerate}
	Finally, with $\lambda^* = \kappa > 0$, from \eqref{E:KKT5} we have $ \TcommOpt = \SumN \tau_n^*$. 	
\end{document}